\documentclass[sigconf]{acmart}
\usepackage{bbold}

\usepackage{amssymb}
\usepackage{amsmath}
\usepackage{algorithm}
\usepackage{algpseudocode}

\usepackage{subfigure}
\usepackage{epstopdf}
\newtheorem{definition}{Definition}

\begin{document}
	
	\title{TFROM: A Two-sided Fairness-Aware Recommendation Model for Both Customers and Providers}
	
	\author{Yao Wu}
	\email{wuyaoericyy@sjtu.edu.cn}
	\author{Jian Cao}
	\email{cao-jian@sjtu.edu.cn}
	\authornote{Corresponding author}
	\affiliation{%
		\institution{Shanghai Jiao Tong University}
		\city{Shanghai}
		\country{China}
	}
	
	\author{Guandong Xu}
	\email{Guandong.Xu@uts.edu.au}
	\affiliation{%
		\institution{University of Technology Sydney}
		\city{Sydney}
		\state{New South Wales}
		\country{Australia}
	}
	
	\author{Yudong Tan}
	\email{ydtan@ctrip.com}
	\affiliation{%
		\institution{Department of Air Ticket Business Ctrip.com International Ltd Shanghai}
		\city{Shanghai}
		\country{China}
	}
	
	\begin{abstract}
		At present, most research on the fairness of recommender systems is conducted either from the perspective of customers or from the perspective of product (or service) providers. However, such a practice ignores the fact that when fairness is guaranteed to one side, the fairness and rights of the other side are likely to reduce. In this paper, we consider recommendation scenarios from the perspective of two sides (customers and providers). From the perspective of providers, we consider the fairness of the providers' exposure in recommender system. For customers, we consider the fairness of the reduced quality of recommendation results due to the introduction of fairness measures. We theoretically analyzed the relationship between recommendation quality, customers fairness, and provider fairness, and design a two-sided fairness-aware recommendation model (TFROM) for both customers and providers. Specifically, we design two versions of TFROM for offline and online recommendation. The effectiveness of the model is verified on three real-world data sets. The experimental results show that TFROM provides better two-sided fairness while still maintaining a higher level of personalization than the baseline algorithms.
	\end{abstract}
	
	
	\maketitle
	
	\section{Introduction}
	At present, most recommender systems aim to maximize the customers' utility by learning their behavior and recommending items that best match their preferences. Research on customer behaviors \cite{senecal2004influence} has proven that recommendations can indeed influence their decisions and result in a good customer experience. However, recommender systems can also bring unfavorable consequences, such as they may narrow the customers' vision \cite{adomavicius2011improving}, or superior items will receive increased attention so as to become dominant \cite{ovaisi2020correcting}, while inferior items will be relegated to a lower position, which becomes an extremely vicious circle. As a possible unfavorable consequence, the unfairness in recommender systems in different aspects, such as racial/gender stereotypes \cite{kay2015unequal}, social polarization \cite{dandekar2013biased}, position bias \cite{ovaisi2020correcting}, has been a well-studied research topic.
	
	\textbf{Problem Statement.} Despite the different mechanisms which have been implemented to ensure the fairness of recommendations, these studies only consider the utility of one type of stakeholder in business and try to eliminate unfairness among their members. This makes sense on a platform where one side dominates. For example, employers have an absolute say in job hunting, and liminating inequality among employees may not harm the interests of employers. However, in most situations, there are multiple stakeholder types on a platform. If the interests of one side are enhanced, the interests of the other side will be damaged. Research has found that when the fairness of customers' recommendation quality is guaranteed, the exposure of providers will be greatly unfair \cite{patro2020fairrec}. Since on most platforms, customers and product providers are the two most important stakeholder types, in this paper, we consider the problem of fair recommendation for these two types of stakeholders. 
	
	The interests of providers in a recommender system are mainly reflected in the positions of their products on the customers' recommendation lists. Providers want their products to be ranked as high as possible since products in higher positions can attract more attention, which in turn brings more orders and higher revenues. But maintaining fairness among providers is necessary to maintain a healthy market environment, which is beneficial to the long-term development of the platform.
	
	Customers would like to receive recommendations from the platform that meet their personal requirements, which is the main goal of recommendation algorithm design. However, due to the introduction of the fairness measurement, customer satisfaction may be lowered because the recommendation lists are not generated by only considering their preferences. It is a challenging task to maintain the fairness of recommendation and also maximize the level of personalization at the same time.
	
	\textbf{State-of-the-art and Limitations.} At present, only a few studies consider two-sided fairness in recommender system, among which \cite{patro2020fairrec} is one representative. Although \cite{patro2020fairrec} also discusses fairness between customers and providers, it is limited in that one provider only corresponds to one item so the problem becomes assuring fairness both for customers and the exposure of items. It does not conform to the reality that a provider often offers multiple items, which is what we are trying to solve in this paper. It is also worth noting that one provider offering one item is a special case of our problem. In addition, \cite{patro2020fairrec} also ignores the impact of the positions of items in the recommendation list on item exposure, and regards all the items appearing in a list as gaining the same exposure rate. However, according to the research \cite{joachims2007search}, the top ranked items receive more attention. Therefore, we introduce the metrics considering the influence of position on exposure.
	
	Many papers classify fairness in recommender systems into individual fairness \cite{rastegarpanah2019fighting} and group fairness \cite{asudeh2019designing}. Individual fairness emphasizes the similar treatment of similar individual users, while group fairness emphasizes that the benefits of the group or the probability of receiving services accord with the demographic structure. In fact, both approaches are reasonable, and their differences stem from their objectives. Since most of the literature only considers fairness for one type of stakeholder, it is natural for them to either consider individual fairness or group fairness. In this paper, we try to ensure recommendation fairness both for providers and consumers so that individual fairness among customers and group fairness among providers are both considered.
	
	\textbf{Approach and Contribution.} When two-sided fairness is considered, recommendation becomes a multi-objective problem. Since these objectives conflict with each other, it is impossible to optimize each objective at the same time, but it is possible to find a relatively better trade-off among multiple objectives through algorithms. We theoretically analyze the relationship between recommendation quality, customer fairness, and provider fairness, and find the direction of problem optimization. We design two algorithms for online and offline application scenarios. In the offline scenario, we generate recommendation lists for all customers. In the online scenario, customers' requests arrive randomly and recommendation lists are generated for each request. 
	
	The contributions of this paper are as follows:
	\begin{itemize}
		\item We propose and formulate the problem of two-sided fairness in recommender systems. By considering the influence of product position in a recommendation list, we design new metrics to measure the \textit{individual fairness} of customers, \textit{group fairness} of providers and the quality of the recommendation results.
		\item  Through theoretical analysis, we design TFROM (a Two-sided Fairness-aware RecOmmendation Model), which is implemented in two versions for online and offline scenarios. TFROM can be easily applied to various existing recommender systems.
		\item The experiment results on three real-world datasets show that TFROM can provide better two-sided fairness and still maintain a higher recommendation quality than the comparison algorithms.
	\end{itemize}
	
	The rest of the paper is organized as follows. Section 2 discusses the related work. Section 3 formalizes the two-sided fairness problem. Section 4 presents the TFROM. The experiment results are detailed in Section 5. We conclude the paper in Section 6.

	\section{Related Work}
	With the increasing maturity of recommendation technology, many researchers have begun to focus on metrics other than recommendation accuracy to measure the performance of recommender systems \cite{ge2010beyond,pu2011user}, and fairness is one of the important metrics.
	
	According to stakeholders considered in the algorithm, research on fairness in recommender systems can be divided into the following three categories \cite{burke2017multisided}: those that consider customer-side fairness \cite{sonboli2020opportunistic,bobadilla2020deepfair}, those that consider provider-side fairness \cite{zehlike2020reducing,lei2020multi,morik2020controlling} and those that consider two-sided fairness \cite{chakraborty2017fair}. Research which considers consumer-side fairness usually aims at eliminating discrimination suffered by some customers in the recommendation process and enables different customers to have the same experience. For example, a re-ranking algorithm is proposed in \cite{geyik2019fairness} that mitigates the bias of protected attributes such as gender or age. In \cite{bose2019compositional}, information on protected sensitive attributes is removed in graph embedding by learning a series of adversarial filters. Provider-side fairness usually focuses on providing a fair channel for different providers to reach their customers. For example, in \cite{biega2018equity}, an amortization algorithm is designed that allows recommended items to gain exposure commensurate with their quality. In \cite{qian2015scram}, a fair taxi route recommendation system is proposed so that taxi drivers can have a fair chance of accessing passengers.
	
	While the vast majority of relevant work considers only unilateral  stakeholder fairness, we consider fairness from both sides. When both stakeholder sides are considered, there will be more objectives in the algorithm, and the fairness for the two sides may be in conflict with each other, bringing new problems which need to be solved. \cite{suhr2019two} discusses two-sided fairness in a ride-hailing platform to ensure fairness in driver income and user waiting time. \cite{wang2020fairness} discusses the relationship between user fairness, item fairness and diversity in intent-aware ranking. In \cite{patro2020fairrec}, an algorithm is designed based on a greedy strategy to ensure providers receive fair exposure and customers receive fair recommendation quality, which is consistent with the goal of our research. However, in the setting of \cite{patro2020fairrec}, each item is treated as a provider, which is a special case in our research problem. In our model, we assume each provider can provide one or more items, which is more in line with the application scenario in real life. 
	
	There are also some taxonomies that classify the fairness of recommendations from other perspectives. In some studies, the fairness of recommendations is divided into individual fairness \cite{rastegarpanah2019fighting,boratto2020connecting} and group fairness \cite{asudeh2019designing,beutel2019fairness,ge2021towards}. Group fairness is intended to eliminate the influence of specific attributes on the recommendation results for different groups so that disadvantaged groups are offered the same opportunities as the advantaged groups, whereas the goal of individual fairness is to enable similar users to be treated similarly. Approaches can also be classified from the perspective of the time that the mechanism works in the system \cite{zafar2017fairness}, and the fairness mechanism is divided into pre-processing \cite{calmon2017optimized}, in-processing \cite{agarwal2018reductions,bose2019compositional,wu2020fairness} and post-processing \cite{liu2018personalizing,karako2018using} approaches. Our study considers both individual and group fairness. Provider-side fairness focuses on the fairness of the groups of items provided by each provider, while from the perspective of customers we focus on fairness between individual customers. We propose a post-processing approach that further processes the existing recommendation results to obtain results that ensure two-sided fairness.
	
	\section{Problem Formulation}\label{Problem Formulation}
	In this paper, we consider a recommendation system with two types of stakeholders. The provider is the party who provides the recommended items or services, and each provider can provide one or more items or services. The customer is the one who receives the recommended result.
	
	We assume that there is a recommendation algorithm in the system, which provides a predicted rating matrix $V$ and the original recommendation lists $L^{ori}$ for all customers based on $V$. Our algorithm uses the obtained preference matrix to provide customers with recommendation results that are both in line with their preferences and ensure two-sided fairness.
	
	\subsection{Notations}
	We use the following notations:
	\begin{itemize}
		\item $U = \{u_1, u_2,..., u_m\}$ is a set of customers.
		\item $I = \{i_1, i_2,..., i_n\}$ is a set of recommended items.
		\item $P = \{p_1, p_2,..., p_l\}$ is a set of providers supplying items. \item $I_p$ is the set of items provided by provider $p$.
		\item $V = \left[v_{u_1,i_1}, v_{u_1,i_2},..., v_{u_m,i_n}\right]$ is a relevant rating matrix produced by the original recommendation algorithm of the system.
		\item $L^{ori} = \{\textbf{l}^{ori}_{u_1}, \textbf{l}^{ori}_{u_2},..., \textbf{l}^{ori}_{u_m}\}$ is a set of original recommendation lists based on $V$.
		\item $L = \{\textbf{l}_{u_1}, \textbf{l}_{u_2},..., \textbf{l}_{u_m}\}$ is a set of recommendation lists finally outputted to customers.
	\end{itemize}
	
	\subsection{Exposure of items and providers}\label{Exposure of items and providers}
	As previously mentioned, the provider's interest in the recommendation system is reflected in the exposure of its items on the recommendation list, and the exposure of an item depends on its position on the customers' recommendation lists. According to research on user behavior, only top-ranked items tend to attract more attention before the user makes a decision \cite{joachims2007evaluating}, and even an item in position 5 is largely ignored \cite{joachims2007search}. 
	
	To address this phenomenon, we need to give lower weights to lower-ranked items, and this loss of exposure changes very quickly at the beginning as the rank drops, whereas for lower ranked items, as the attention has been reduced, the changes in exposure tend to stabilize. Therefore, the exposure of an item $i$ can be defined as:
	\begin{equation}\label{E1}
		e_i = \sum_{u\in U}\frac{\mathbb{1}_{\textbf{l}_u(i)}}{\log_{2}(r_{u,i}+1)} 
	\end{equation}
	where $\mathbb{1}_{\textbf{l}_u(i)}$ equals $1$ when $i$ is in $\textbf{l}_u$, and $0$ otherwise. The symbol $r_{u,i}$ represents the position of item $i$ in $l_u$.

	A provider's total exposure can be viewed as an aggregation of the exposures of items it provides as $e_p = \sum_{i\in I_p}e_i$.
	
	\subsection{Fairness in Providers' Exposure}
	There are different definitions of exposure fairness in relevant studies. In some definitions, better items should have higher exposure \cite{biega2018equity}, that is, it is fair if the exposure of an item is proportional to its quality. Some consider a more universal measure of fairness that gives all items the same exposure. In this paper, we consider these two kinds of exposure fairness at the same time, and our algorithm  can support the optimization of these two kinds of fairness.
	
	Since a provider's exposure is the aggregation of its items' exposures, the provider who offers more items intuitively gains higher exposure, then the fairness between providers can be essentially transformed into the fairness between groups of recommended items. In line with the aforementioned idea of the two kinds of exposure fairness, we have the following definitions:
	
	\begin{definition}[Uniform Fairness]
		A recommendation result has the property of uniform fair exposure for providers if each provider receives exposure proportional to the number of items it offers.
		\begin{equation}\label{E2}
			\frac{e_{p_1}}{|I_{p_1}|} = \frac{e_{p_2}}{|I_{p_2}|},\forall p_1, p_2\in P.
		\end{equation}
	\end{definition}
	
	\begin{definition}[Quality Weighted Fairness]
		A recommendation result is quality weighted fair exposure for providers if each provider receives exposure proportional to the sum of quality scores of items it offers in the recommendation lists of all customers.
		\begin{equation}\label{E2}
			\frac{e_{p_1}}{\sum_{i\in I_{p1}}\sum_{u\in U}v_{u,i}} = \frac{e_{p_2}}{\sum_{i\in I_{p2}}\sum_{u\in U}v_{u,i}},\forall p_1, p_2\in P.
		\end{equation}
	\end{definition}
	
	Fairness of exposure can then be measured in terms of the dispersion of data, such as the variance of exposure, and the lower the degree of dispersion of the data, the more fairness it indicates.
	
	\subsection{Quality of Recommendation}\label{Quality of Recommendation}
	The introduction of a fairness index into recommendation results will reduce the quality of recommendation results since some items with low ratings will be allocated to higher positions in the list in order to ensure fair exposure for providers. In this paper, we define the quality of the recommendation results as the degree to which the recommendation list matches the original list of the recommender system. It is worth pointing out that although the original recommendation list may still not be absolutely in line with customers' preferences in practice, it can be assumed that the original recommendation list has already reflected their preferences as much as possible. 
	
	We use two classic metrics in information retrieval, namely discounted cumulative gain (DCG) and normalized discounted cumulative gain (NDCG) to measure the quality of recommendation \cite{biega2018equity}. DCG sums up the relative scores of all items in the recommended list and gives a logarithmic discount based on their ranks which is consistent with the idea of item exposure. 
	
	\begin{equation}
		DCG_{u,\textbf{l}} = v_{u,\textbf{l}[1]} + \sum_{i=2}^k\frac{v_{u,\textbf{l}[i]}}{\log_{2}(i+1)}
	\end{equation}
	
	NDCG further normalizes DCG by dividing the DCG value of a customer's original recommendation list $\textbf{l}_u^{ori}$ (IDCG) which is the ideal situation in terms of recommendation quality for results completely being in line with customer preferences. 
	
	\begin{equation}
		NDCG_{u} = \frac{DCG_{u,\textbf{l}_u}}{DCG_{u,\textbf{l}^{ori}_u}}
	\end{equation} 
	
	By dividing by the DCG of the original recommendation list, the difference in customers' scoring habits can be eliminated. When this value is equal to 1, it indicates that the result is fully in line with the customer's preference, and the smaller the value, the greater the loss of recommendation quality.         
	
	\subsection{Measuring the Fairness of Recommendation for Customers}
	As previously mentioned, when the fair exposure of providers is taken into consideration, the quality of recommendations will be reduced. We introduce the idea of \textbf{individual fairness} and want the recommendation quality reduction to be equally allocated to every customer. We provide the following definition: 
	
	\begin{definition}[Fair recommendation for customers]
		The recommendation is fair for customers if each customer receives recommendation results with the same NDCG value.
		\begin{equation}\label{E2}
			NDCG_{u_1} = NDCG_{u_2},\forall u_1, u_2\in U.
		\end{equation}
	\end{definition}
	
	We can also use variance of NDCG values of customers' recommendation results to measure the overall customer-side fairness.
	
	\subsection{Trade-off between Customer Benefits and Provider Benefits}
	In this section, we discuss the relationship between customer recommendation quality, customer fairness, and provider exposure fairness, and clarify the direction of our algorithm optimization.
	
	As demonstrated experimentally in \cite{patro2020fairrec}, if the recommendation system is completely subordinate to customer preferences, it can result in vastly unfair provider exposure. When the algorithm tries to modify the original recommendation list to improve the fairness of the providers' exposure, we get the following three theorems, which describe the relationships between the three objectives:
	
	\textbf{THEOREM 1.} \emph{There is no such an algorithm that does not reduce the quality of recommendation results compared with the original list $\textbf{l}_u^{ori}$, unless the algorithm directly recommends the original recommendation list to customers.}
	
	\textbf{THEOREM 2.} \emph{When the recommendation quality decreases from the best situation, the fairness of customer recommendation quality decreases or remains unchanged.}
	
	\textbf{THEOREM 3.} \emph{When the fairness of provider exposure increases, the fairness of customer recommendation quality decreases or remains unchanged.}
	
	As mentioned in Section \ref{Quality of Recommendation}, if the algorithm directly recommends the original recommendation list $\textbf{l}_u^{ori}$ to customers, the recommendation quality of customers is at the best level (NDCG is equal to 1), and the quality among customers is fair. Since the original recommendation list is the optimal case of the customers' recommendation quality, \textbf{THEOREM 1} must be correct. 
	
	When the recommendation quality decreases, because the distribution of customers recommendation quality is not the same, there will be differences in the degree of quality loss between customers, which increases the variance of customer recommendation quality, and the customer-side fairness decreases. But at the same time, there may be such a special case that the degradation of quality is evenly distributed to all customers, so that the quality of customer recommendations remains the same, and the customer-side fairness is maintained, which is \textbf{THEOREM 2}. Although this special case does not necessarily exist in all data, we can distribute the loss of quality to all customers as much as possible when optimizing provider exposure, so as to make the decrease in customer-side fairness as small as possible.
	
	Combining \textbf{THEOREM 1} and \textbf{THEOREM 2}, we can get the relationship between the fairness of exposure and the fairness of recommendation quality, which is \textbf{THEOREM 3}. This shows that we can improve the fairness of exposure while still maintaining the fairness of recommendation quality. 
	
	Based on these three theorems, we can get the optimization direction of the problem. Regarding the unfairness problem of the providers' exposure when customers’ preferences are fully satisfied, we can sacrifice part of the recommendation quality to optimize the fairness of exposure. At the same time, we can make the quality loss distributed to all customer as evenly as possible to maintain the fairness of customer recommendation quality. The design of our algorithms is based on this idea.
	
	\section{A Two-sided Fairness-aware Recommendation Model (TFROM)}\label{TFROM}
	The two-sided fairness problem discussed in Section \ref{Problem Formulation} can be reduced to a knapsack problem which has been proven to be a non-deterministic polynomial complete problem. We analogize the length of the recommendation list as the capacity of the knapsack, the items to be recommended as the items put in the knapsack, and fairness as the objective. When further taking the quality of recommendation lists into consideration, the problem becomes more complicated. So we choose heuristic strategies to solve the problem. In this section, we propose TFROM to solve the two-sided fairness problem. TFROM consists of two  algorithms designed for two scenarios. One is an offline scenario in which the system makes recommendation to all customers once at the same time, such as via an advertising push. The other is an online scenario where customers' requests arrive randomly, and the system needs to respond to each request within a short period of time and provide the recommendation results, for example, for online purchases.
	
	\subsection{TFROM for Offline Scenario}
	Recommendations can be generated for all customers in an offline fashion, such as advertising via email. In this situation, 
	we need to select $k$ items from $n$ items for $m$ customers respectively. The exposure brought by the position of each item in the recommendation list can be calculated. If the length of the recommendation list $k$ and the total number of customers $m$ are known, the total exposure provided by a recommendation can be calculated as follows:
	\begin{equation}
		E_{total} = m\times \sum_{rank=1}^{k}\frac{1}{log_2(rank+1)}
	\end{equation}
	
	Furthermore, based on the previous definitions of exposure fairness, we can calculate how much exposure each provider should obtain to reach a fair state based on the number of items provided by this provider as follows:
	\begin{equation}
		e_{p_l}^{Fair} = \frac{E_{total} \times |I_{p_l}|}{\sum_{p\in P}|I_p|}\text{ (Uniform Fairness)}
	\end{equation}
	\begin{equation}\label{Quality Weighted Fairness}
		e_{p_l}^{Fair} = \frac{E_{total} \times \sum_{i\in I_{p_l}}\sum_{u \in U}v_{u,i}}{\sum_{p\in P}\sum_{i\in I_p}\sum_{u \in U}v_{u,i}}\text{ (Quality Weighted Fairness)}
	\end{equation}
	
	If all providers receive exposure equal to $e_{p_l}^{Fair}$ in a recommendation, then absolute exposure fairness is achieved between providers. Because only a fixed exposure value can be provided by the position in the recommendation list, the total exposure received by each provider is virtually impossible to be equal to the ideal value. But we can use these values as a benchmark for the exposure of each provider, and make the actual exposure as close to these benchmarks as possible, so as to ensure the fairness of exposure between providers as much as possible.
	
	In order to ensure customer fairness, we propose the following approach. In order to distribute the recommendation quality reduction as evenly as possible, customers who have experienced a lower loss in quality in the recommendation process should suffer more losses than customers who have experienced a higher loss in quality at present; in order to improve the overall recommendation quality for customers, it is necessary to give priority to items which are more relevant to customers as much as possible.
	
	Combining the aforementioned ideas, \textit{TFROM-offline} works as follows. The algorithm recommends a list of items from position $1$ to $k$. When making a recommendation for a certain position, the algorithm must wait until all customers are recommended an item in this position before items in the next position can be recommended. For the first position, an item is selected for each customer according to the results of a recommendation algorithm in an arbitrary order. For each of the remaining positions, the customers will be sorted from high to low according to recommendation quality scores in terms of the items selected for them, and the rest items will be selected for them in this order. For each selection, the algorithm finds the highest-ranked and un-recommended item from a customer's original recommendation list $\textbf{l}_u^{ori}$. If the item's provider's previous exposure plus the item’s exposure at this position does not exceed the provider’s fair exposure baseline $e_p^{Fair}$, the item is selected, otherwise the algorithm will look for the next item along the original recommendation list $\textbf{l}_u^{ori}$ until a suitable item is found. If all the items in the $\textbf{l}_u^{ori}$ list do not meet the conditions, then this position will be skipped and will be allocated after all positions have been tentatively filled with items.
	
	Every time an item is selected, the exposure of the provider $e_p$ and the recommendation quality obtained by the customers $NDCG_u$ are updated. This process is repeated until items in position $k$ have been selected. After this, the positions that are skipped before are re-allocated from high positions to low positions. Each time, \textit{TFROM-offline} recommends an unrecommended item with the lowest provider exposure in the $\textbf{l}_u^{ori}$ list to further reduce the difference in exposure between providers and ensure that all positions are filled. The pseudo-code of \textit{TFROM-offline} is shown in \textbf{Algorithm 1}.
	
	\begin{algorithm}
		\label{alg1}
		\caption{Two-sided Fairness-aware Recommendation Model for Offline Scenario}
		\begin{algorithmic}[1]
			\Require
			$k$: The number of items recommended for each customer;\newline 
			$\textbf{l}_{u_1}^{ori}$, $\textbf{l}_{u_2}^{ori}$,..., $\textbf{l}_{u_m}^{ori}$: Original recommendation list of $m$ customers;\newline
			\textbf{V}: Rating matrix;\newline
			$I_{p_1}$, $I_{p_2}$,...,$I_{p_l}$: The set of items provided by $l$ providers;\newline
			$e_{p_1}^{Fair}$, $e_{p_2}^{Fair}$,..., $e_{p_l}^{Fair}$: Fair exposure of each provider;
			\Ensure
			$\textbf{l}_{u_1}$, $\textbf{l}_{u_2}$,..., $\textbf{l}_{u_m}$: Recommendation results for customers;
			\State $Q = [q_{u_1}, q_{u_1},..., q_{u_m}] \gets [0 \times m]$; //recommendation quality
			\State $e_{p_1}$, $e_{p_2}$,..., $e_{p_l} \gets 0$; // exposure of providers
			\State $\textbf{l}_1$, $\textbf{l}_2$,..., $\textbf{l}_m \gets [-1 \times k]$;
			\State $\textbf{l}_{u_1}^{un\_rec}$, $\textbf{l}_{u_2}^{un\_rec}$,..., $\textbf{l}_{u_m}^{un\_rec} \gets \textbf{l}_{u_1}^{ori}$, $\textbf{l}_{u_2}^{ori}$,..., $\textbf{l}_{u_m}^{ori}$;
			\For{$rank = 1 \to k$}
			\If{rank == 1}
			\State $Sorted\_customer \gets \text{ }$Random order.
			\Else
			\State $Sorted\_customer \gets \text{ }$Sort customers according to $rec\_q_u$ from the highest to lowest.
			\EndIf 
			\For{$u\_temp$ in $Sorted\_customer$}
			\For{$i\_temp$ in $l_{u\_temp}^{un\_rec}$}
			\State $p\_temp = i\_temp.provider$
			\If{$e_{p\_temp} + \frac{1}{log_2(rank+1)} \leqslant e_{p\_temp}^{Fair}$}
			\State $\textbf{l}_{u\_temp}[rank] = item\_temp$;
			\State $e_{p\_temp} += \frac{1}{log_2(rank+1)}$;
			\State $q_{u\_temp} += \frac{v_{u\_temp,i\_temp}}{log_2(rank+1)\times IDCG_{u\_temp}}$;
			\State $\textbf{l}_{u\_temp}^{un\_rec}.remove(i\_temp)$ ;
			\State break;
			\EndIf
			\EndFor
			\EndFor
			\EndFor
			\For{$rank = 1 \to k$}
			\For{$u\_temp$ in $U$}
			\If{$\textbf{l}_{u\_temp}[rank] == -1$}
			\State $e_{min} = Inf$;
			\For{$i\_temp$ in $l_{u\_temp}^{un\_rec}$}
			\If{$e_{i\_temp.provider} <= e_{min}$}
			\State $i\_next = i\_temp$;
			\EndIf
			\EndFor
			\State $\textbf{l}_{u\_temp}[rank] = i\_next$;
			\State $e_{i\_next.provider} += \frac{1}{log_2(rank+1)}$;
			\State $q_{u\_temp} += \frac{v_{u\_temp,i\_next}}{log_2(rank+1)\times IDCG_{u\_temp}}$;
			\State $\textbf{l}_{u\_temp}^{un\_rec}.remove(i\_next)$;
			\EndIf
			\EndFor
			\EndFor
			\State\Return $\textbf{l}_{u_1}$, $\textbf{l}_{u_2}$,..., $\textbf{l}_{u_m}$; 
		\end{algorithmic}	
	\end{algorithm}

	\subsection{TFROM for Online Scenario}\label{Online}
	In an online situation, customer requests arrive randomly and the algorithm needs to respond to each request in a timely manner. In this case, the fairness of a single round of recommendation loses its meaning, and our target is changed into long-term fairness. We transform the customer's recommendation quality and the provider's exposure into a cumulative value. In addition, due to the different number of times the recommendation services are provided for each customer, the recommendation quality for each customer should be further divided by the number of times the recommendations are provided, which is the average of the recommendation quality obtained by each customer in the service process. 
	
	At the same time, because the number of customer requests constantly increases, the total exposure cannot be determined in advance as in the offline situation, but changes dynamically with the arrival of customer requests. The fair exposure of providers thus needs to be recalculated as the total exposure changes:
	
	\begin{equation}
		\begin{split}
			&E_R = c\_num\times \sum_{rank=1}^{k}\frac{1}{log_2(rank+1)}\\
			&\textit{where } c\_num \textit{ is the number of customer requests.}
		\end{split}	
	\end{equation} 
	
	If the algorithm continues to follow the idea of \textit{TFROM-offline} and reduces the exposure difference between providers by only adjusting this customer's recommendation list, the quality of recommendations for him may be greatly reduced. Therefore, when filling vacancies in the recommendation list, we give priority to recommendation quality and directly select the remaining items with the highest preference scores in the $\textbf{l}_u^{ori}$ list.
	
	At the same time, when the number of customer requests is very small, the calculated fair exposure baseline of each provider is smaller than the exposure that will be provided by the recommendation, which will result in no items being selected in the early stage. When filling vacancies, \textit{TFROM-online} selects items in the order of preference scores, and in this case, it happens to be directly recommending the $\textbf{l}_u^{ori}$ list to customers, which is also a very reasonable strategy in the initial stage of system startup. This shows that \textit{TFROM-online} is suitable for the inception phase, and can naturally transition to the regular stage without additional operations. The pseudo-code of \textit{TFROM-online} is shown in \textbf{Algorithm 2}.
	
	\begin{algorithm}
		\label{alg2}
		\caption{Two-sided Fairness-aware Recommendation Model for Online Scenario}
		\begin{algorithmic}[1]
			\Require
			$k$: The number of items recommended for each customer;\newline
			$u$: The coming customer;\newline
			$\textbf{l}_{u}^{ori}$: Original recommendation list of the coming customer $u$;\newline
			\textbf{V}: Rating matrix;\newline
			$I_{p_1}$, $I_{p_2}$,...,$I_{p_l}$: The set of items provided by $l$ providers;\newline
			$e_{p_1}^{Fair}$, $e_{p_2}^{Fair}$,..., $e_{p_l}^{Fair}$: Fair exposure of each provider;\newline
			$e_{p_1}$, $e_{p_2}$,..., $e_{p_l}$: Accumulated Exposure of $l$ providers up to last recommendation;\newline
			$q_u$: Average recommendation quality of customers $u$ up to last recommendation;\newline
			$rec\_time_u$: The number of times that the recommendation services customer $u$ has received up to last time recommendation;
			\Ensure
			$\textbf{l}_u$: Recommendation results for the coming customer $u$;
			\State $\textbf{l}_u\gets [-1 \times k]$;
			\State $\textbf{l}_u^{un\_rec}\gets l_u^{ori}$;
			\State $q_u^{temp} = 0$;
			\For{$rank = 1 \to k$}
			\For{$i\_temp$ in $l_u^{un\_rec}$}
			\State $p\_temp = i\_temp.provider$;
			\If{$e_{p\_temp} + \frac{1}{log_2(rank+1)} \leqslant e_{p\_temp}^{Fair}$}
			\State $\textbf{l}_u[rank] = item\_temp$;
			\State $e_{p\_temp} += \frac{1}{log_2(rank+1)}$;
			\State $q_u^{temp} += \frac{v_{u,i\_temp}}{log_2(rank+1)\times IDCG_u}$;
			\State $\textbf{l}_u^{un\_rec}.remove(i\_temp)$;
			\State break;
			\EndIf
			\EndFor
			\EndFor
			\For{$rank = 1 \to k$}
			\If{$\textbf{l}_u[rank] == -1$}
			\State $i\_next = l_u^{un\_rec}[0]$;
			\State $\textbf{l}_{u\_temp}[rank] = i\_next$;
			\State $e_{i\_next.provider} += \frac{1}{log_2(rank+1)}$;
			\State $q_u^{temp} += \frac{v_{u,i\_next}}{log_2(rank+1)\times IDCG_u}$;
			\State $\textbf{l}_u^{un\_rec}.remove(i\_next)$;
			\EndIf
			\EndFor
			\State $q_u = \frac{q_u\times rec\_time_u + q_u^{temp}}{rec\_time_u + 1}$;
			\State $rec\_time_u += 1$;
			\State\Return $\textbf{l}_u$; 
		\end{algorithmic}	
	\end{algorithm}
	
	\subsection{Time complexity}
	The time complexity of \textit{TFROM-offline} is analyzed as follows. Before selecting items for a certain position, \textit{TFROM-offline} first sorts the customers according to the recommendation quality scores obtained by the items selected for their lists. The complexity of sorting $m$ customers is $O(m\log(m))$ when using the Quick Sort Algorithm or the Merge Sort Algorithm. Then \textit{TFROM-offline} iterates through the original list of customers receiving recommendations to find suitable items that will not exceed the fairness exposure of providers, and in the worst case, the algorithm needs to traverse the list again in the second stage. So in the worst case, the complexity of selecting items for a certain position is $O(2nm^2\log(m))$. Over the whole process, the algorithm needs to select items for $k$ positions, and the worst case time complexity of \textit{TFROM-offline} is $O(knm^2\log(m))$. Since \textit{TFROM-online} only deals with a single customer at a time, its worst case time complexity is $O(2kn)$.

	\section{Experiments}\label{Experiments}
	\subsection{Datasets and Metrics}
	We conducted experiments on three datasets - a flight dataset from an online travel company Ctrip \footnote{https://www.ctrip.com}, a Google local dataset and an Amazon review dataset. These three data sets cover three very important aspects of a customer's daily life, i.e., travel, local living and online shopping. Our code and datasets are released on Zenodo \footnote{https://zenodo.org/record/4527725\#.YCMxKegzZPY}.
	
	\subsubsection{Ctrip Flight Dataset}
	We select the ticket order data on a popular international flight route from Shanghai to Seoul from 2017 to 2020, and treat tickets from the same airline, of the same class and in the same departure time as the same item, and the airline to which the ticket belongs as the provider. The entire dataset contains data of 3,814 customers, 6,006 kinds of air tickets and 25,190 orders, and it provides basic information on customers, as well as air ticket price, air ticket class, airline company of the ticket, flight time and other ticket information. We adopt the state-of-the-art collaborative filtering air ticket recommendation algorithm  \cite{gu2019addressing} to process the data, and obtain the original recommendation lists and customer-item preference matrix.
	
	\subsubsection{Google Local dataset}
	This dataset was released in \cite{pasricha2018translation} and contains reviews about local businesses from Google Maps. We consider businesses located in California, filter out businesses and customers with less than 10 reviews, and obtain a dataset containing 3,335 users, 4,927 businesses, and 97,658 reviews. We mainly consider information such as ratings, comment time, business location, etc., and adopt the state-of-the-art location-based latent factorization algorithm \cite{he2014location} to process the data. The reviews in this dataset are for businesses and do not provide information on the reviewed items. Therefore, for this dataset, we regard each business as a provider, and each provider only provides one item, which is also a special case in our problem.
	
	\subsubsection{Amazon Review dataset}
	This dataset contains a variety of product reviews from Amazon and we use the data released in \cite{he2016ups}. Because of the sheer volume of data, we pre-filter customers and items with less than 10 comments, and only consider reviews of items in the "Clothing Shoes and Jewelry" category, which has the largest number of reviews. We use the state-of-the-art matrix factorization model \cite{liang2016factorization} to obtain the preference matrix. Since the data set does not provide information on the providers of the items, we randomly aggregate 1-100 items to simulate providers with different scales. The processed dataset contains 1,851 users, 7,538 items, 161 providers and 24,658 reviews.
	
	\subsubsection{Metrics}
	We measure the variance of the provider's exposure $e_p$ to evaluate provider-side \textit{Uniform Fairness}, measure the variance of the ratio of provider exposure and relevance in \textit{Equation}(\ref{Quality Weighted Fairness}) to evaluate \textit{Quality Weighted Fairness}. Since the magnitude of the numerator and denominator is quite different, we use the [0,1]-normalized value for calculation.
	
	For the customer-side metrics, we measure the variance of the customer's recommendation quality $NDCG_u$ to evaluate customer-side fairness, and measure the sum of the customers' recommendation quality $NDCG_u$ to evaluate the overall quality of recommendation results. The smaller the variance, the fairer the recommendation results. The greater the sum of $NDCG_u$, the smaller the loss of the recommendation quality.
	
	\begin{figure*}[!h]
		\centering
		\subfigure[Total recommendation quality]{
			\begin{minipage}[t]{0.25\linewidth}
				\centering
				\includegraphics[width=\textwidth,height=2.8cm]{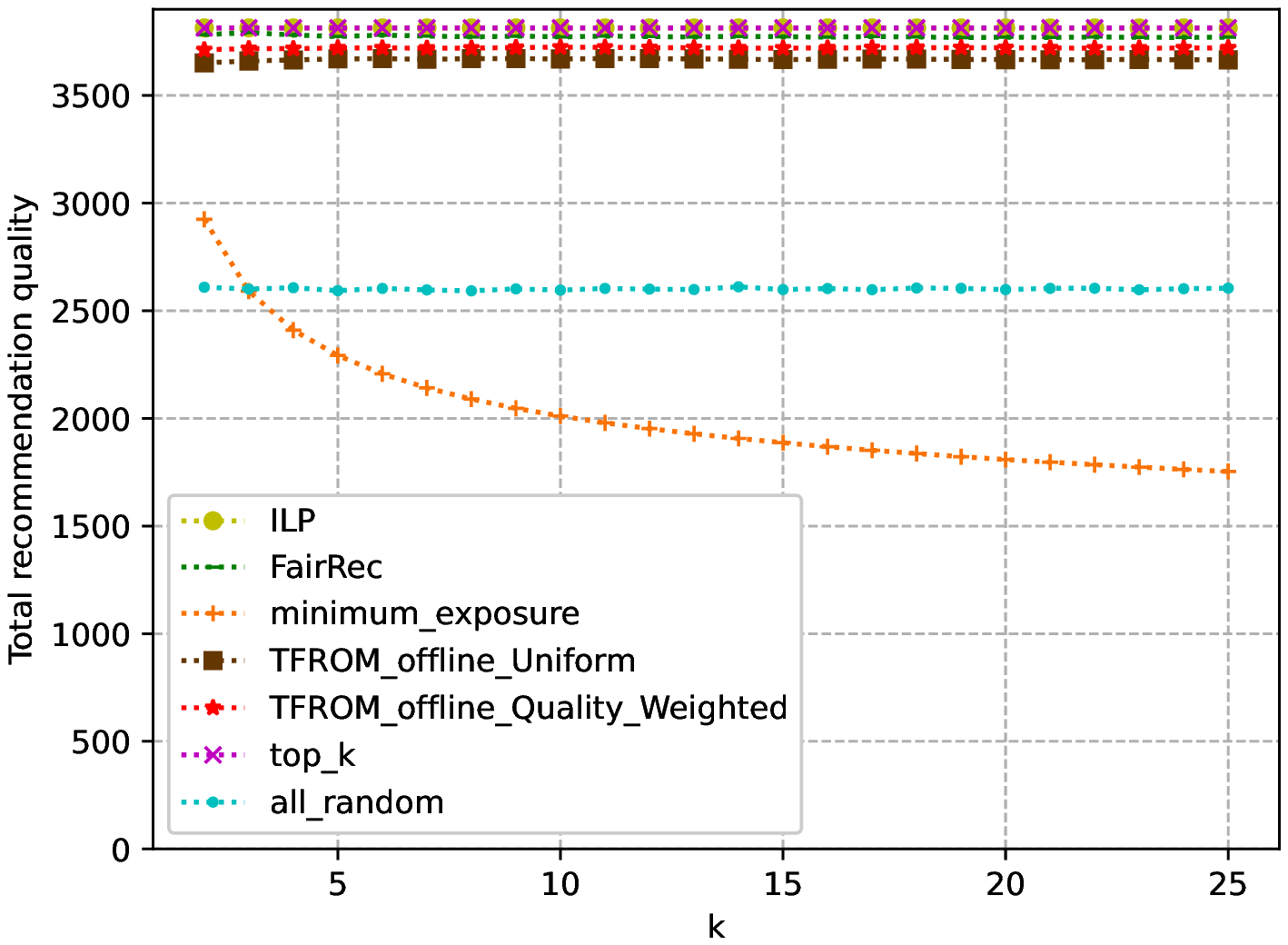}
			\end{minipage}%
		}%
		\subfigure[Variance of NDCG]{
			\begin{minipage}[t]{0.25\linewidth}
				\centering
				\includegraphics[width=\textwidth,height=2.8cm]{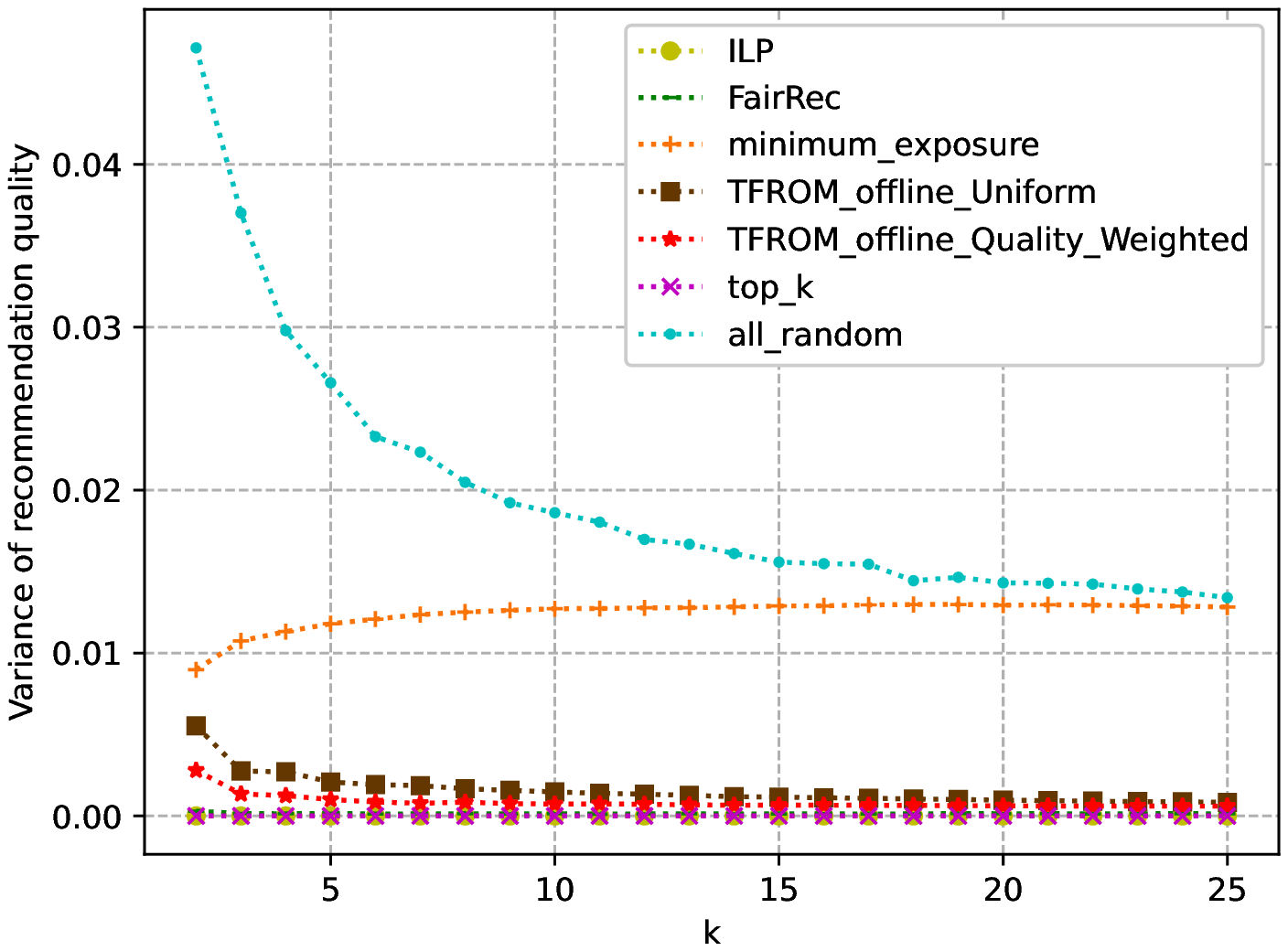}
			\end{minipage}
		}%
		\subfigure[Variance of exposure]{
			\begin{minipage}[t]{0.25\linewidth}
				\centering
				\includegraphics[width=\textwidth,height=2.8cm]{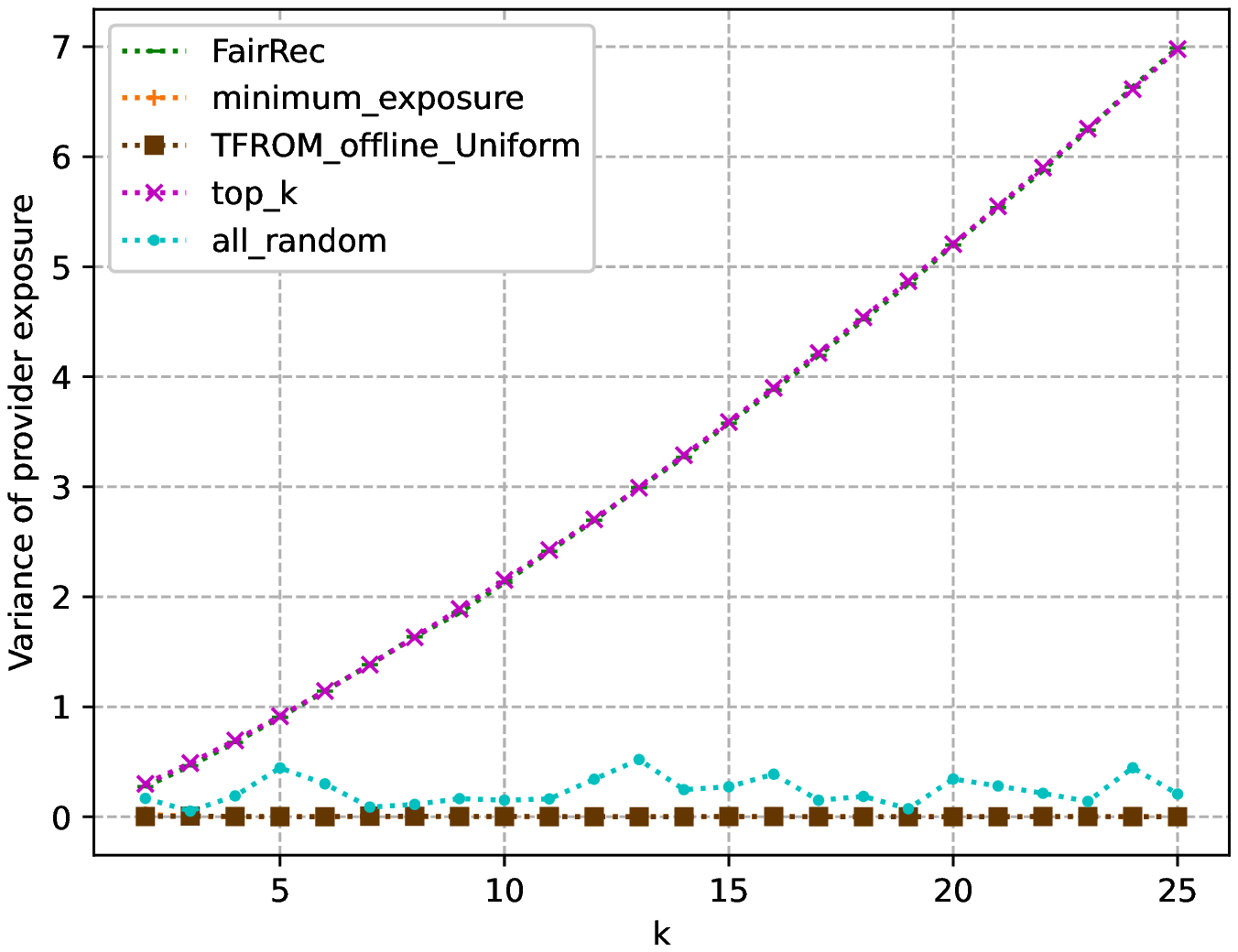}
			\end{minipage}
		}%
		\subfigure[Variance of the ratio of exposure and relevance]{
			\begin{minipage}[t]{0.25\linewidth}
				\centering
				\includegraphics[width=\textwidth,height=2.8cm]{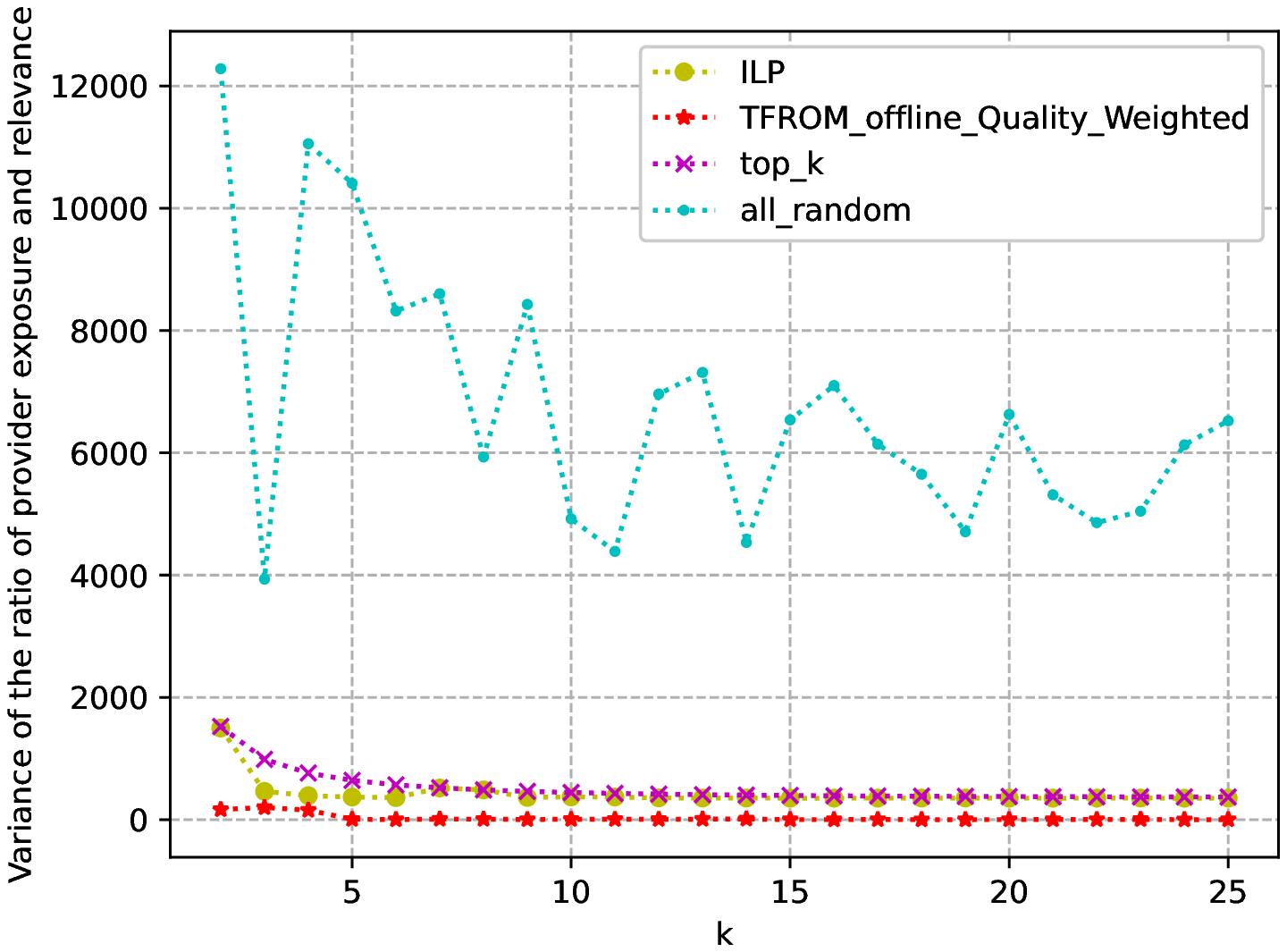}
			\end{minipage}
		}%
		\centering
		\caption{Experiment Results on Ctrip Dataset in the Offline Scenario}
		\label{fig2}
	\end{figure*}

	\begin{figure*}[!h]
		\centering
		\subfigure[Total recommendation quality]{
			\begin{minipage}[t]{0.25\linewidth}
				\centering
				\includegraphics[width=\textwidth,height=2.8cm]{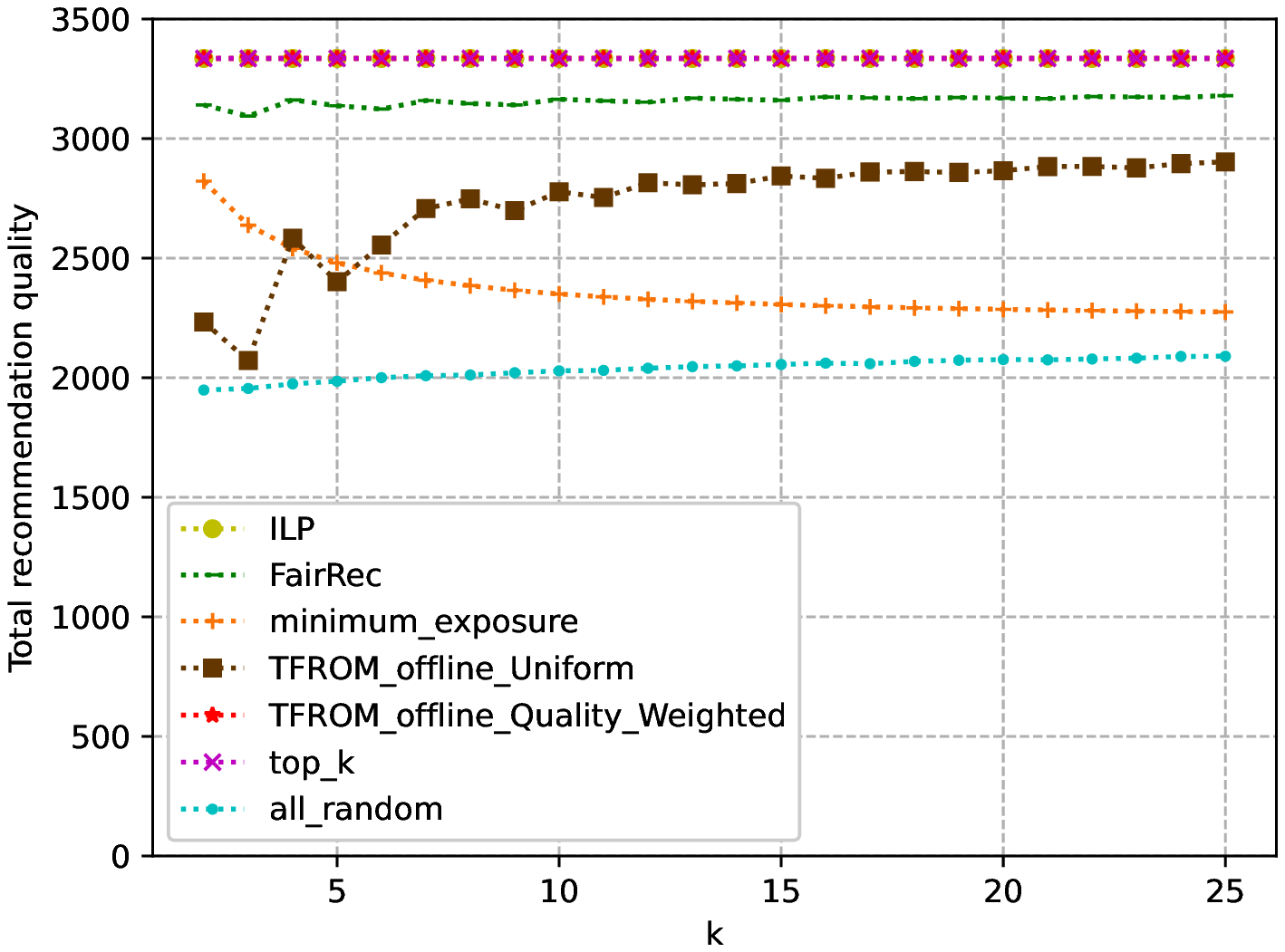}
			\end{minipage}%
		}%
		\subfigure[Variance of NDCG]{
			\begin{minipage}[t]{0.25\linewidth}
				\centering
				\includegraphics[width=\textwidth,height=2.8cm]{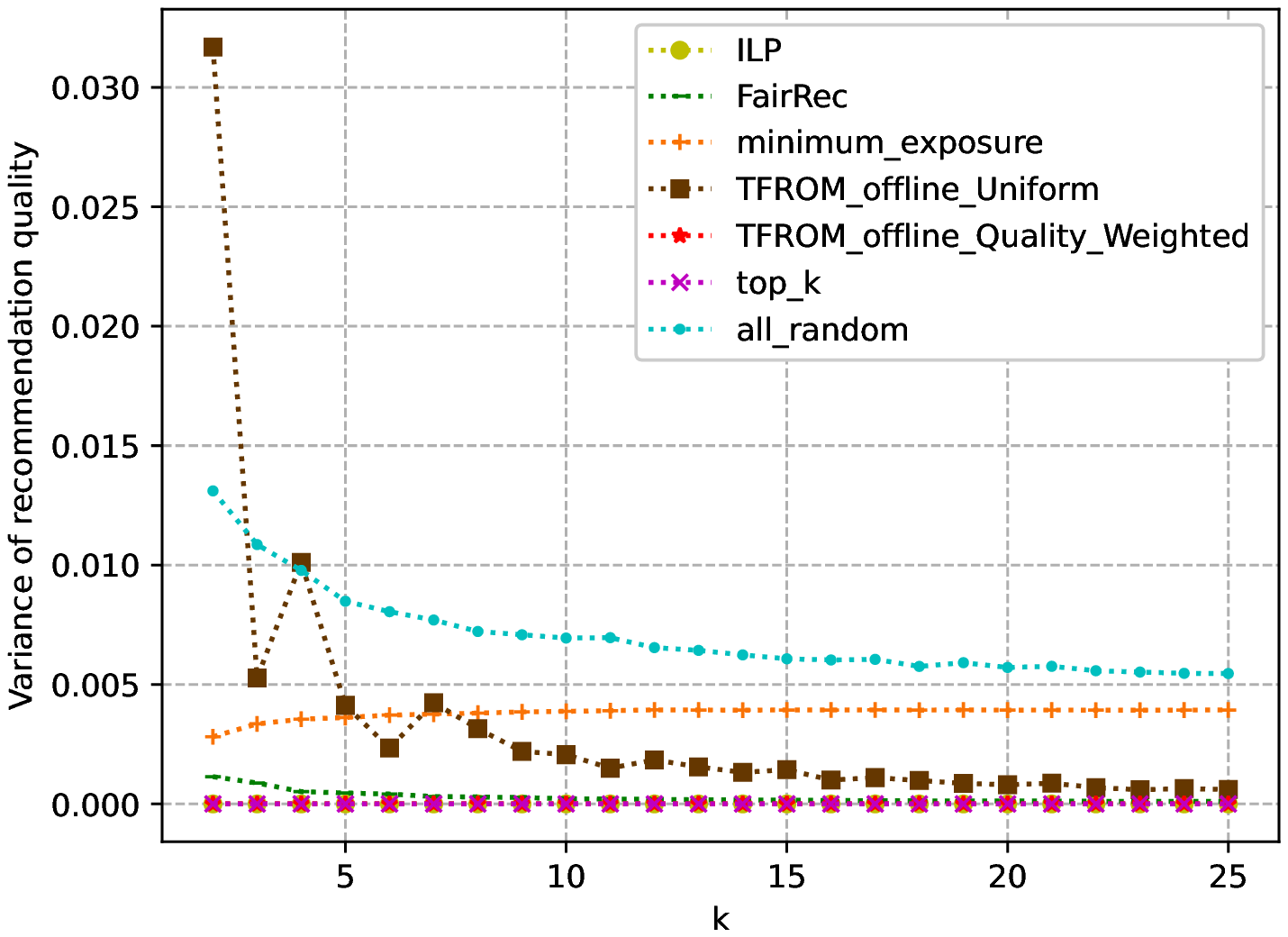}
			\end{minipage}
		}%
		\subfigure[Variance of exposure]{
			\begin{minipage}[t]{0.25\linewidth}
				\centering
				\includegraphics[width=\textwidth,height=2.8cm]{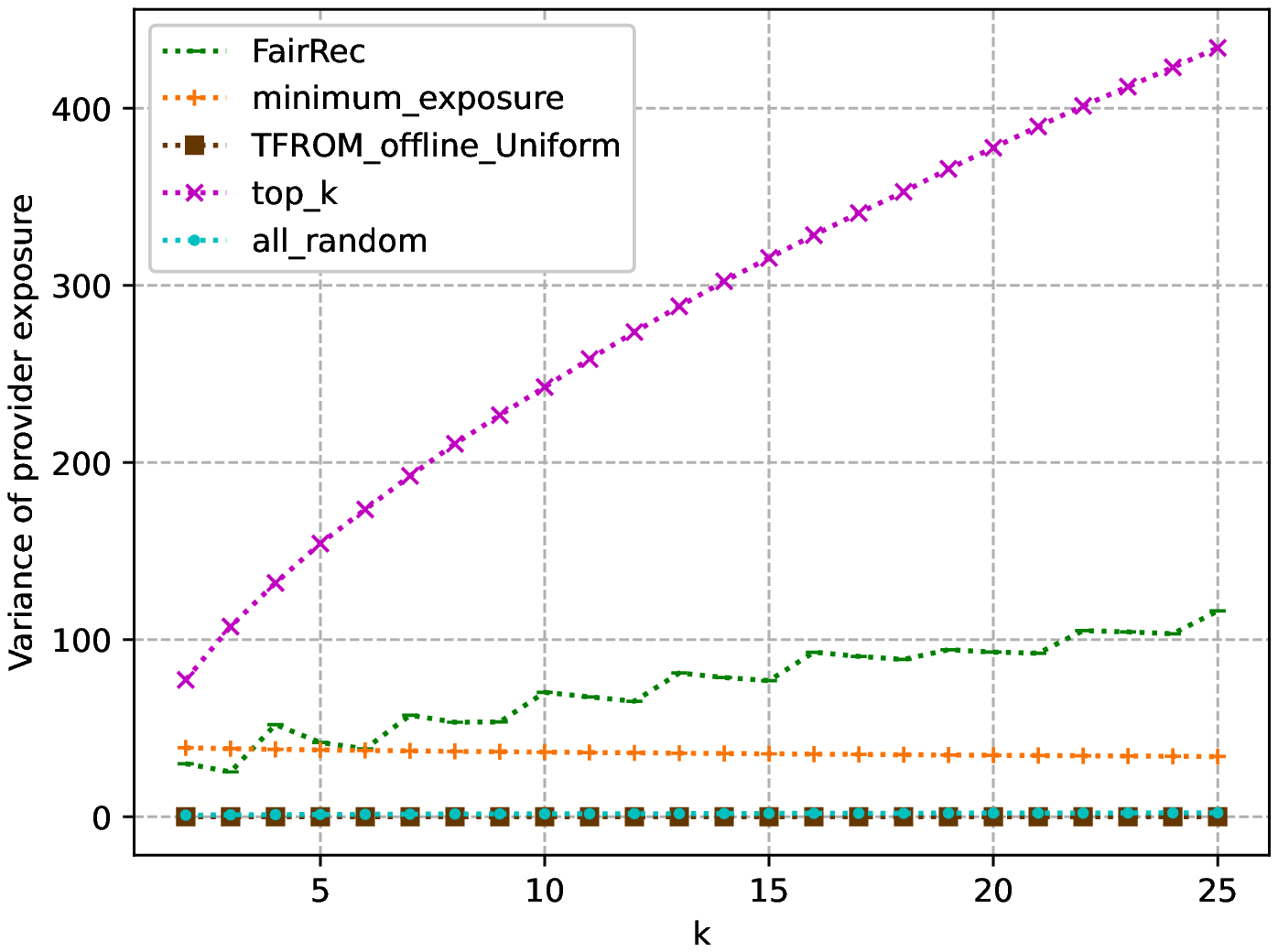}
			\end{minipage}
		}%
		\subfigure[Variance of the ratio of exposure and relevance]{
			\begin{minipage}[t]{0.25\linewidth}
				\centering
				\includegraphics[width=\textwidth,height=2.8cm]{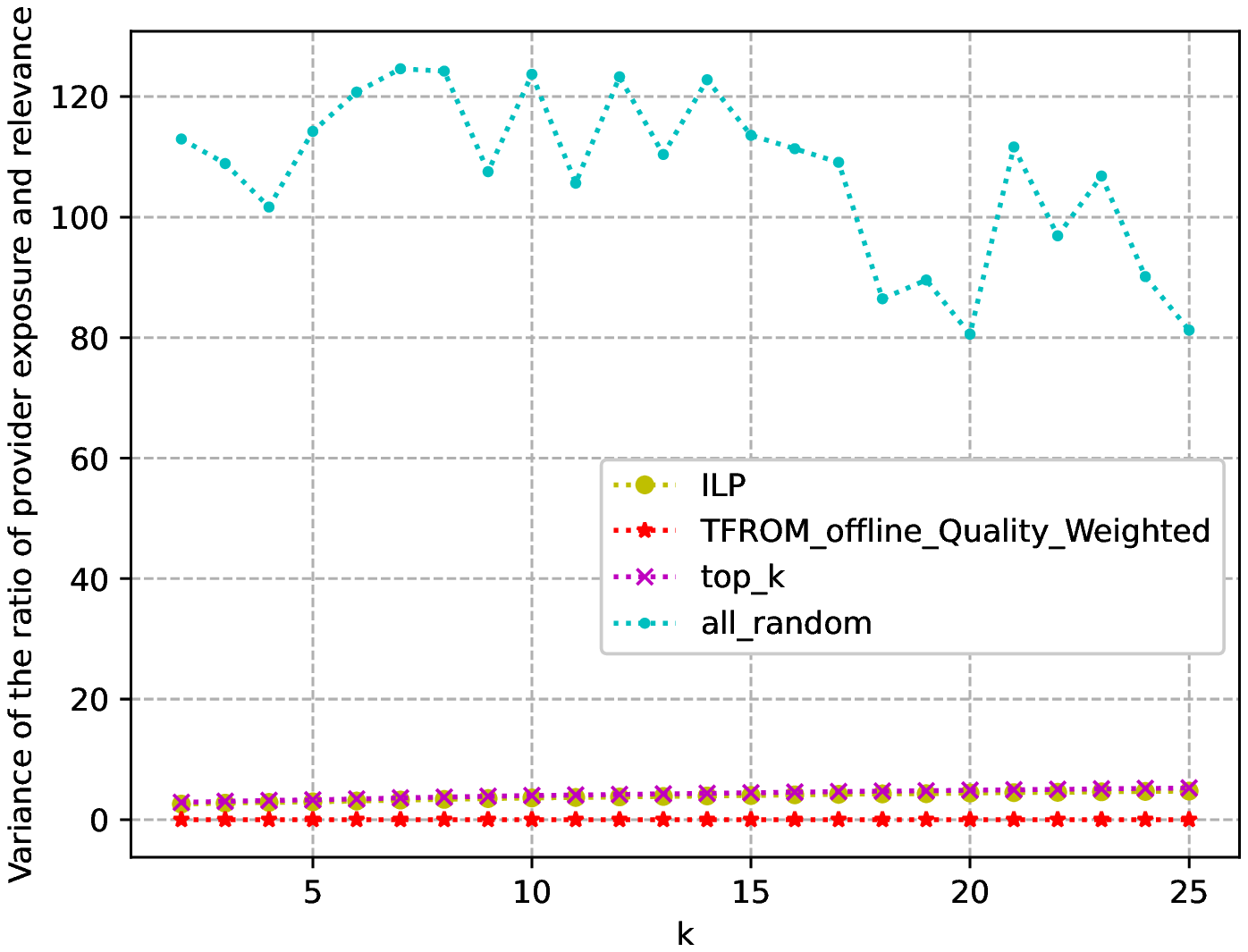}
			\end{minipage}
		}%
		\centering
		\caption{Experiment Results on Google Dataset in the Offline Scenario}
		\label{fig3}
	\end{figure*}
	
	\begin{figure*}[!h]
		\centering
		\subfigure[Total recommendation quality]{
			\begin{minipage}[t]{0.25\linewidth}
				\centering
				\includegraphics[width=\textwidth,height=2.8cm]{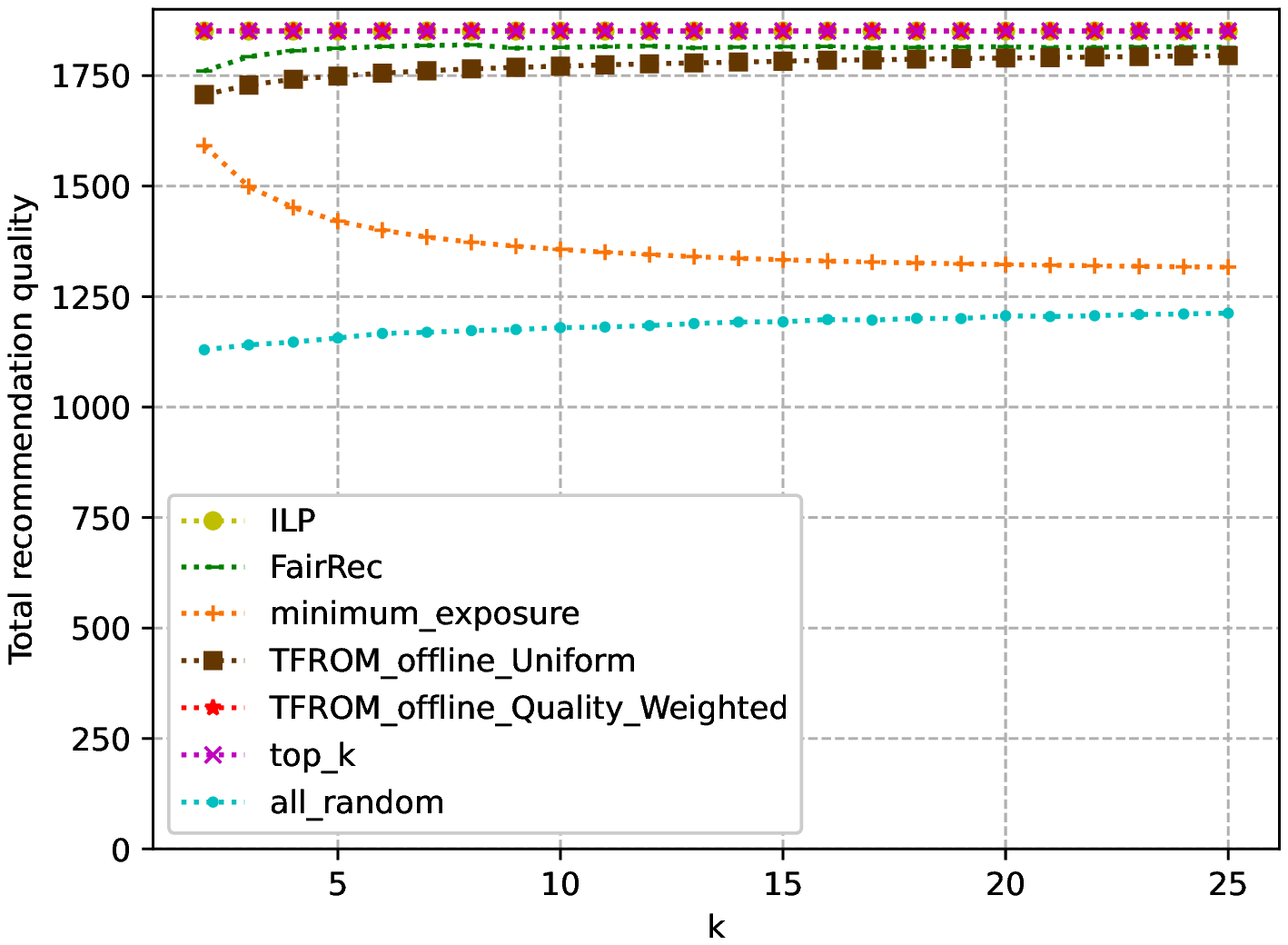}
			\end{minipage}%
		}%
		\subfigure[Variance of NDCG]{
			\begin{minipage}[t]{0.25\linewidth}
				\centering
				\includegraphics[width=\textwidth,height=2.8cm]{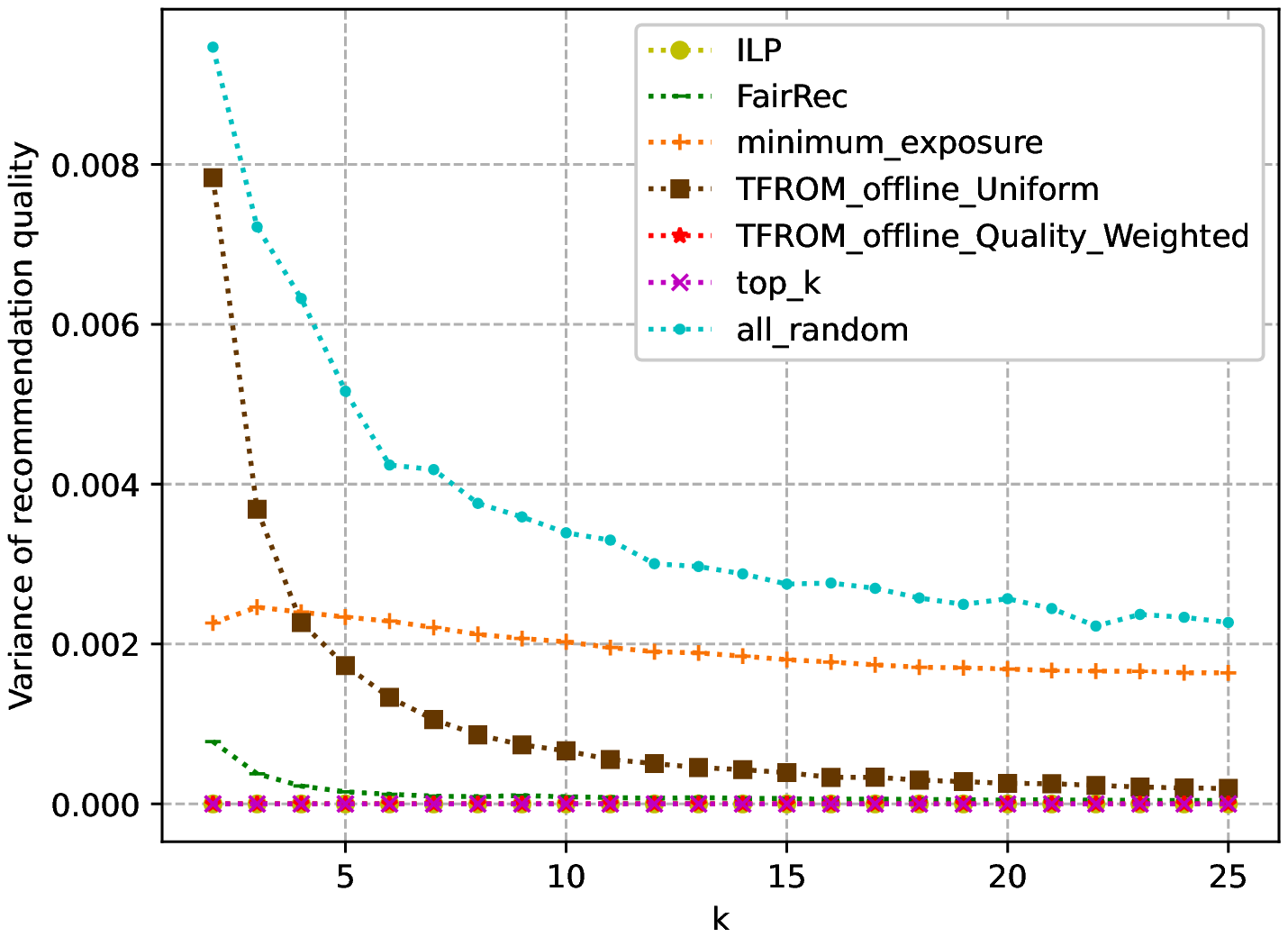}
			\end{minipage}
		}%
		\subfigure[Variance of exposure]{
			\begin{minipage}[t]{0.25\linewidth}
				\centering
				\includegraphics[width=\textwidth,height=2.8cm]{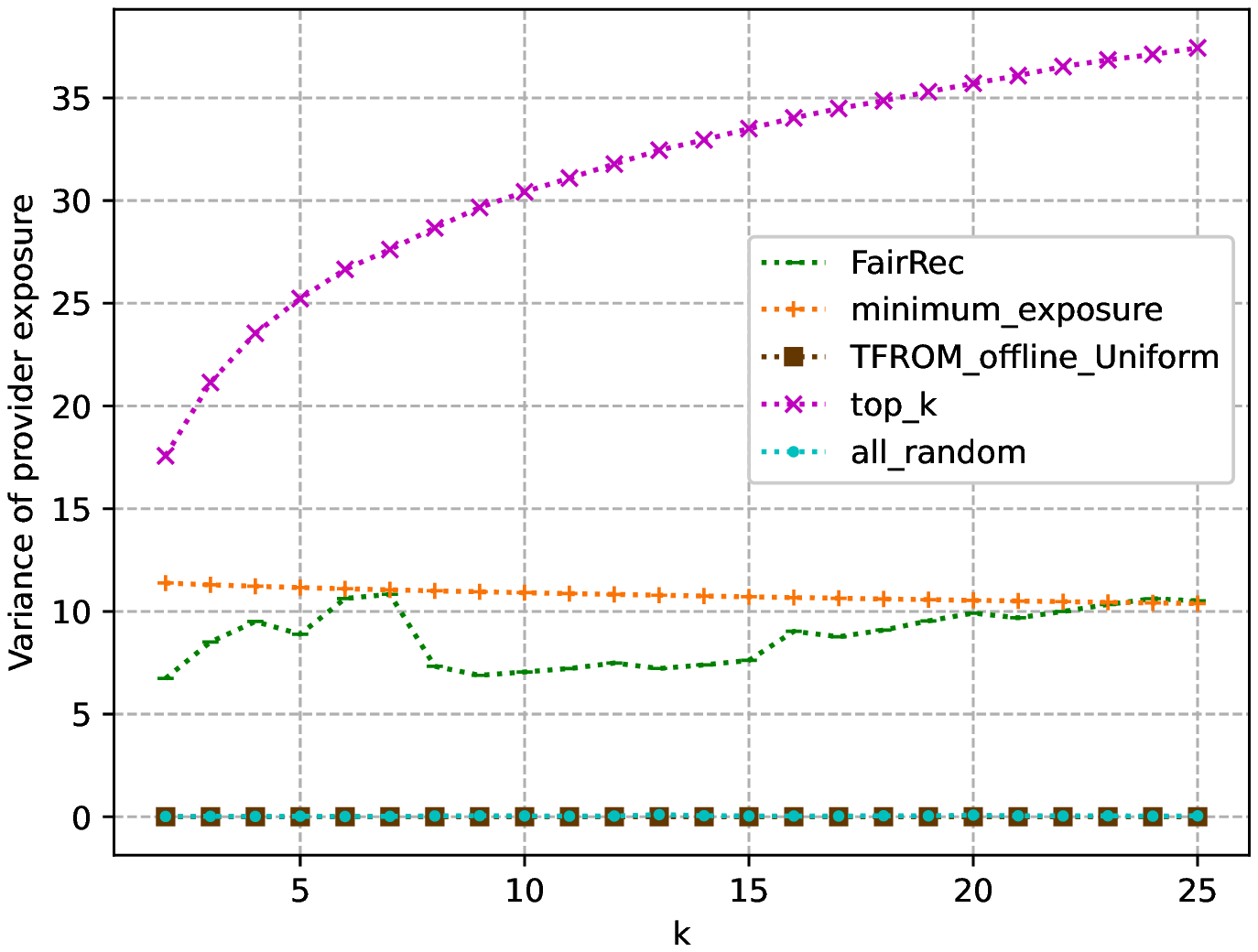}
			\end{minipage}
		}%
		\subfigure[Variance of the ratio of exposure and relevance]{
			\begin{minipage}[t]{0.25\linewidth}
				\centering
				\includegraphics[width=\textwidth,height=2.8cm]{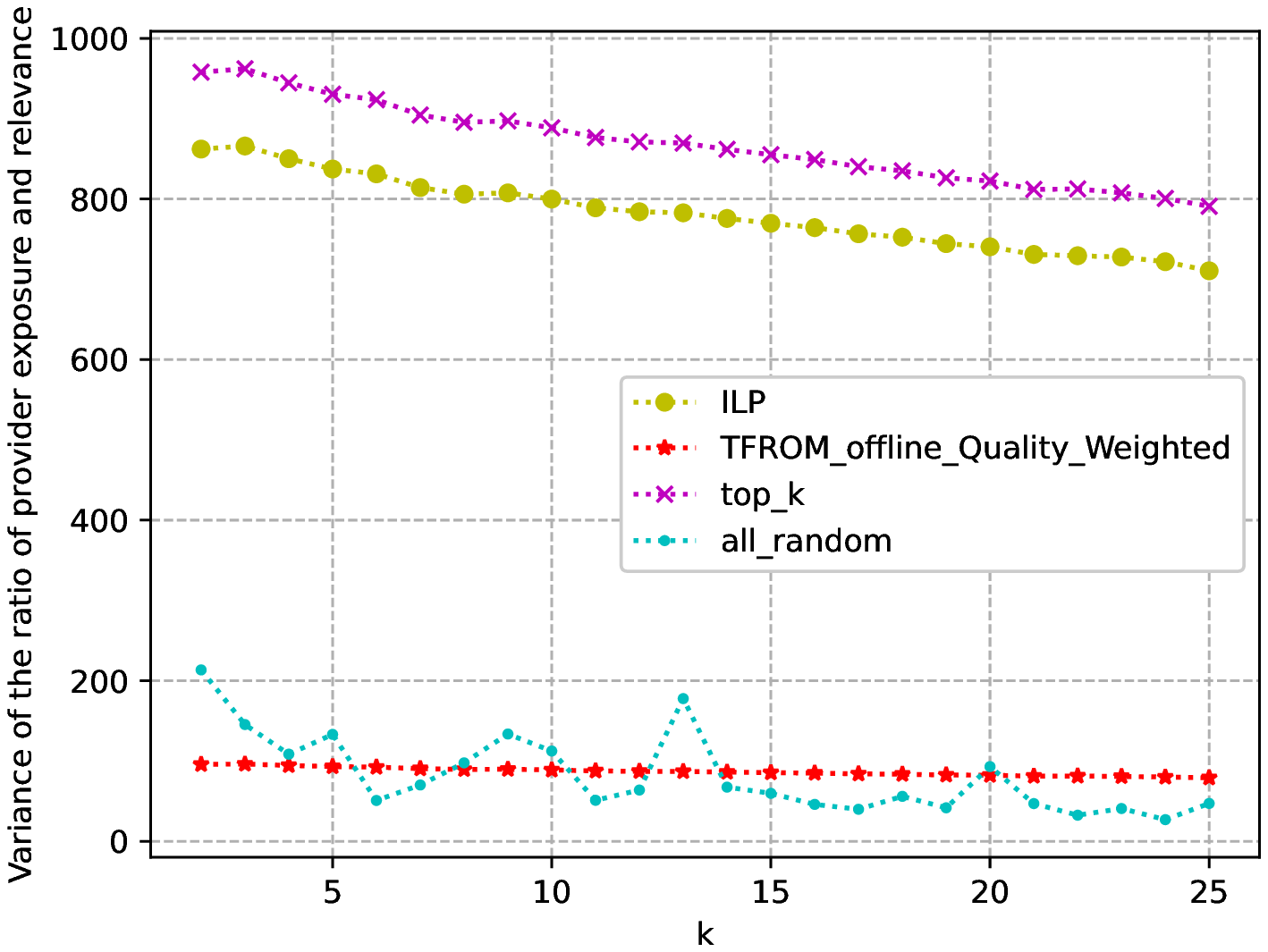}
			\end{minipage}
		}%
		\centering
		\caption{Experiment Results on Amazon Dataset in the Offline Scenario}
		\label{fig4}
	\end{figure*}
	
	\subsection{Compared Approaches}
	We compare our proposed approach with the following algorithms.
	\subsubsection{Top-k}
	This algorithm directly recommends the top-k items in the original recommendation list $\textbf{l}_u^{ori}$, which is also the case for maximizing recommendation quality.
	\subsubsection{All random}
	This algorithm randomly selects $k$ items from the customer's original recommendation list $\textbf{l}_u^{ori}$ to recommend.
	\subsubsection{Minimum exposure}
	This algorithm selects items from the least exposed provider each time for recommendation. This is an algorithm ensure that the providers' exposure is as fair as possible.
	\subsubsection{FairRec}
	This is a state-of-the-art algorithm that guarantees two-sided fairness based on a greedy strategy \cite{patro2020fairrec}, which ensures \textit{Uniform Fairness} for providers by setting the minimum exposure, and fairness for customers using a greedy strategy.
	\subsubsection{An ILP-based fair ranking mechanism}
	This is an algorithm based on integer linear programming(ILP) proposed by \cite{biega2018equity} to ensure the \textit{Quality Weighted Fairness} of provider exposure. This algorithm takes the absolute value of the difference between the two cumulative values as the objective, and the quality of recommendation as the limiting condition.

	\subsection{Experiment Results and Analysis}
	\subsubsection{Results for the offline situation}
	We conducted experiments for the offline situation on three data sets and evaluated the results of the algorithms at different $k$ values.
	
	\paragraph{Recommendation quality}
	As shown in Figure \ref{fig2}(a), \ref{fig3}(a) and \ref{fig4}(a), \textit{Top-k}, \textit{FairRec} and \textit{TFROM-offline-Uniform} produce higher recommendation quality than  \textit{All random} and \textit{Minimum exposure} methods. Of these, \textit{Top-k} achieves the maximum value for recommendation quality as its results are completely in line with the customer's preferences. It is worth noting that the quality loss of \textit{TFROM-offline-Quality-Weighted} is large when $k$ is small, because the algorithm selects items with low relevance when adjusting the exposure fairness. In this case, the cost, i.e., the recommendation quality loss, is large, which results in excessive loss. This situation will be alleviated as $k$ increases. \textit{All random} and \textit{Minimum exposure} algorithm cause a large loss in the recommendation quality due to the lack of special treatment for the recommendation quality. 
	
	In summary, \textit{TFROM-offline} is capable of maintaining an acceptable loss of recommendation quality (less than 10\% in most cases), and in practice, this loss of quality is spread evenly across multiple items so the customer experience will not be changed much and the customers may not even be aware of it. 
	
	\paragraph{Customer-side Fairness}
	The associated results are shown in Figures \ref{fig2}(b), \ref{fig3}(b) and \ref{fig4}(b). It can be seen that all the algorithms provide good customer fairness for all datasets, and the results of \textit{TFROM-offline} are also at a high level in comparison algorithms. In principle, the all random algorithm guarantees fairness for customers, because it carries out the same operation for all customers. However, it only makes a recommendation to each customer once in offline situations, which leads to unsatisfactory results. The \textit{Minimum exposure} algorithm does not operate on customer fairness, so the main reason for the good effect is that there is a large loss in recommendation quality, and as the overall level of recommendation quality is very low, which reduces the difference in customer recommendation quality.
	
	\paragraph{Provider-side Fairness}
	As can be seen from Figures \ref{fig2}(c), \ref{fig3}(c) and \ref{fig4}(c), \textit{TFROM-offline-Uniform}, the minimum exposure algorithm and all random algorithm stably provide fair exposure results on all three datasets as $k$ grows when considering \textit{Uniform Fairness}. From the results, it can also be seen that if customer preferences are completely respected, the inequality of providers will increase with the increase of $k$ and will even increase exponentially. It is worth noting that \textit{FairRec}'s results are not good. Although \textit{FairRec} is designed to ensure fair exposure at the level of individual items, it is not as good as it would have been if multiple items had been aggregated to a provider.
	
	The results of \textit{Quality Weighted Fairness} are shown in \ref{fig2}(d), \ref{fig3}(d) and \ref{fig4}(d). It can be seen that \textit{TFROM-offline-Quality-Weighted} can provide better and more stable fairness on the three datasets. Although the \textit{All random} algorithm can provide better fairness on the Google dataset in some cases, the performance on the other two datasets is not satisfactory. It is worth noting that the results of the \textit{ILP-based method} seem to indicate it has limited optimization capabilities based on the results of \textit{Top-k} algorithm, which may be caused by insufficient solution set space due to pre-filtering.
	
	\begin{figure*}[!h]
		\centering
		\subfigure[Total recommendation quality]{
			\begin{minipage}[t]{0.25\linewidth}
				\centering
				\includegraphics[width=\textwidth,height=2.5cm]{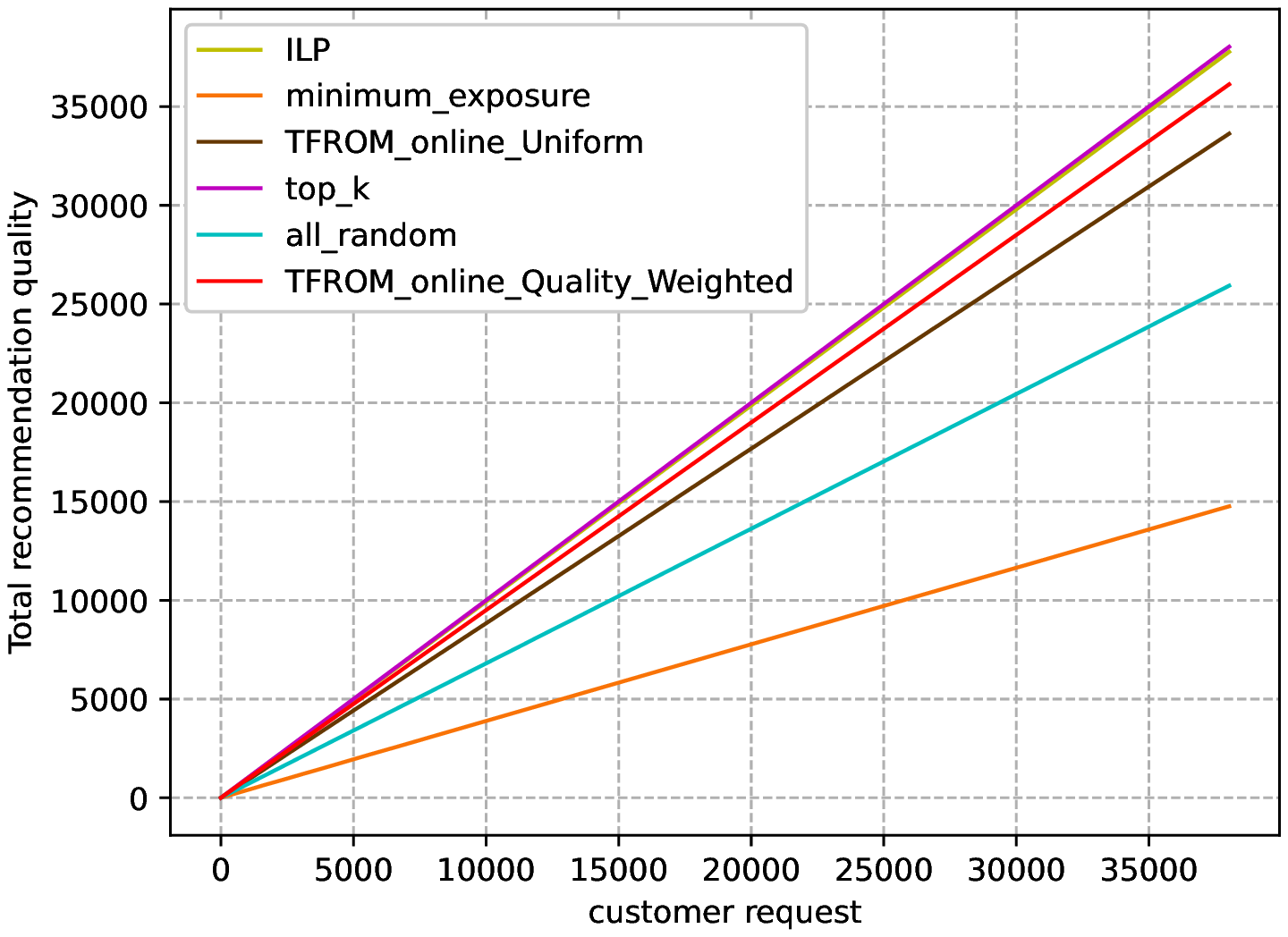}
			\end{minipage}%
		}%
		\subfigure[Variance of NDCG]{
			\begin{minipage}[t]{0.25\linewidth}
				\centering
				\includegraphics[width=\textwidth,height=2.5cm]{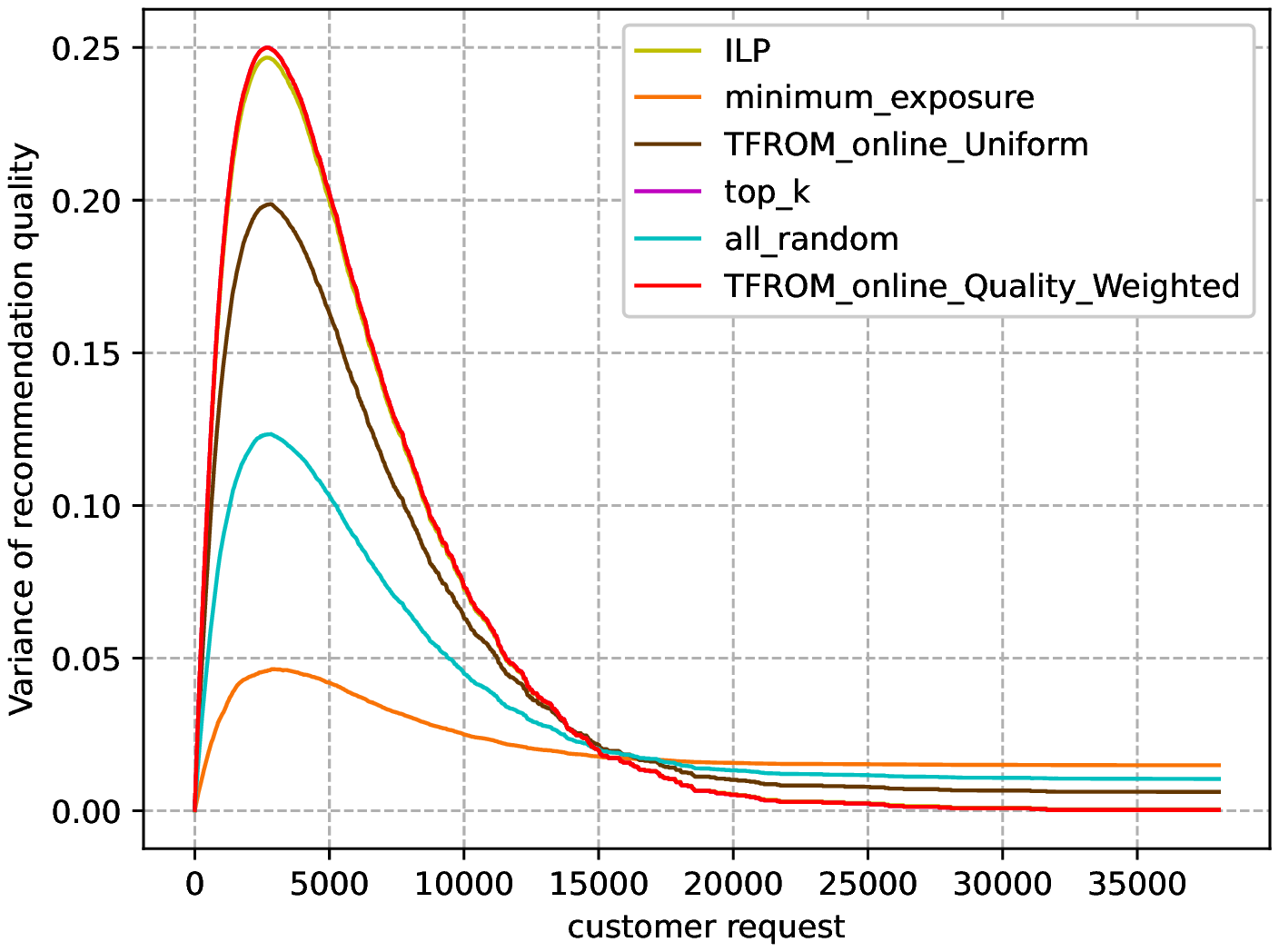}
			\end{minipage}
		}%
		\subfigure[Variance of exposure]{
			\begin{minipage}[t]{0.25\linewidth}
				\centering
				\includegraphics[width=\textwidth,height=2.5cm]{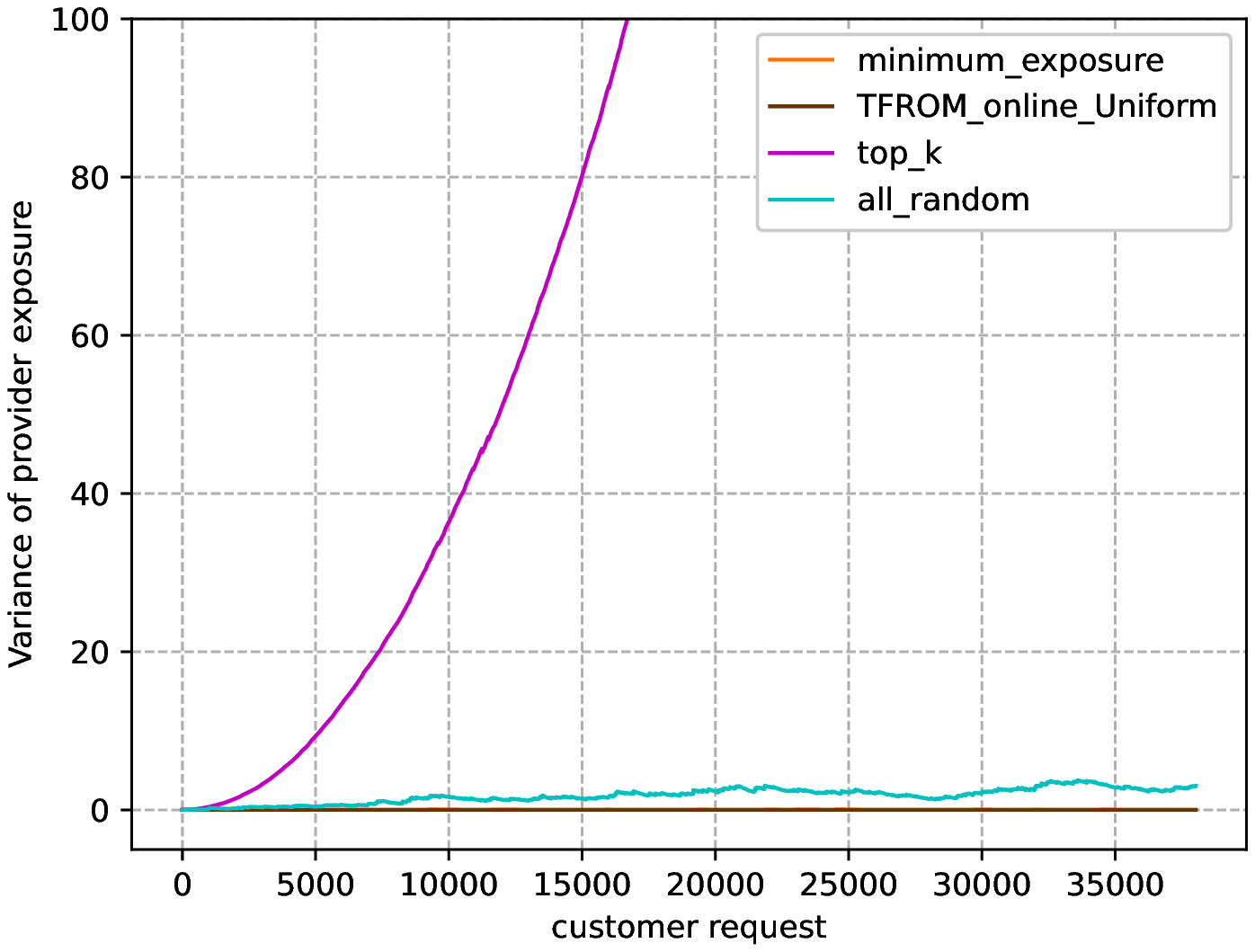}
			\end{minipage}
		}%
		\subfigure[Variance of the ratio of exposure and relevance]{
			\begin{minipage}[t]{0.25\linewidth}
				\centering
				\includegraphics[width=\textwidth,height=2.5cm]{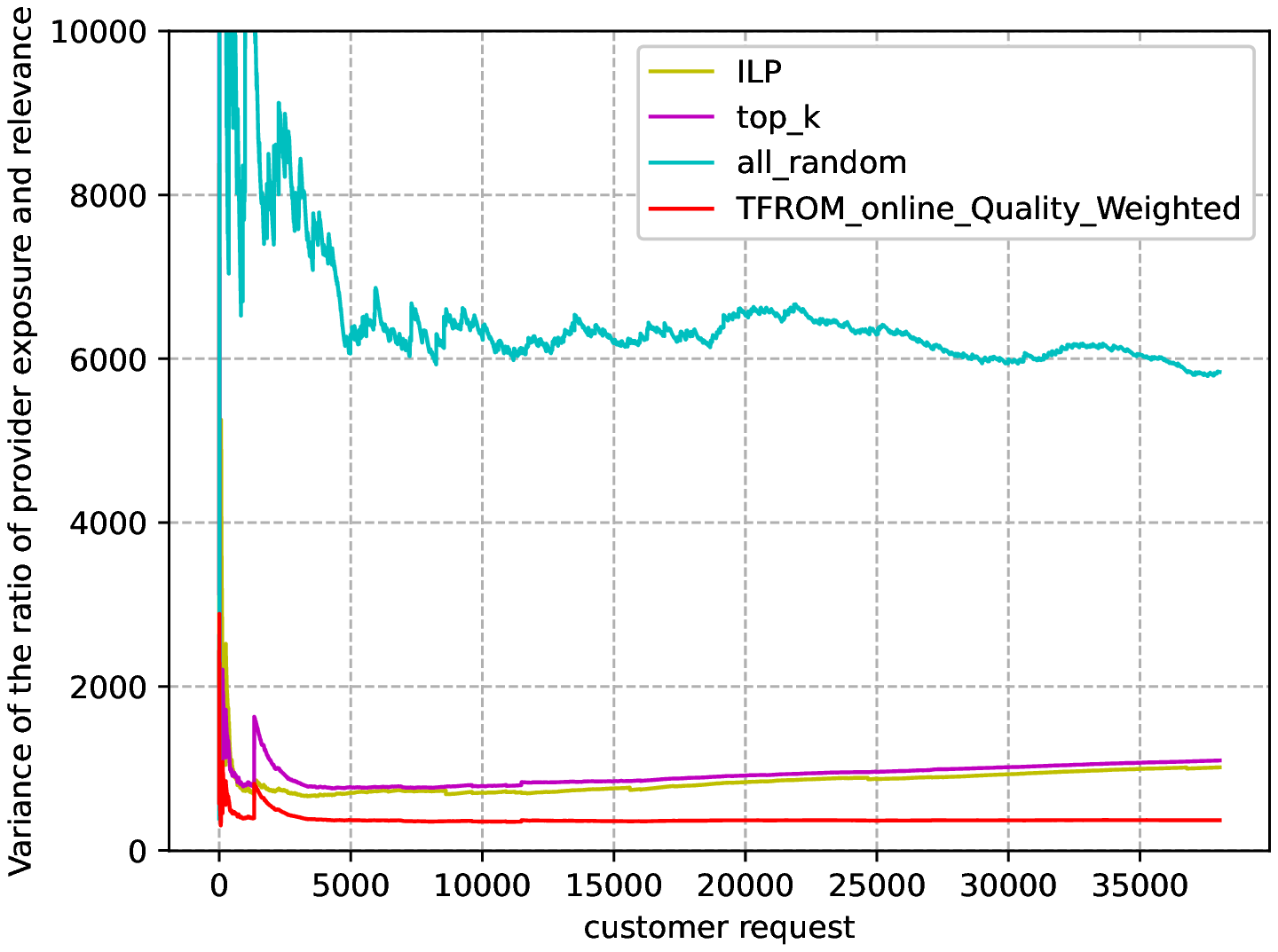}
			\end{minipage}
		}%
		\centering
		\caption{Experiment Results on Ctrip Dataset in the Online Scenario}
		\label{fig5}
	\end{figure*}
	
	\begin{figure*}[!h]
		\centering
		\subfigure[Total recommendation quality]{
			\begin{minipage}[t]{0.25\linewidth}
				\centering
				\includegraphics[width=\textwidth,height=2.5cm]{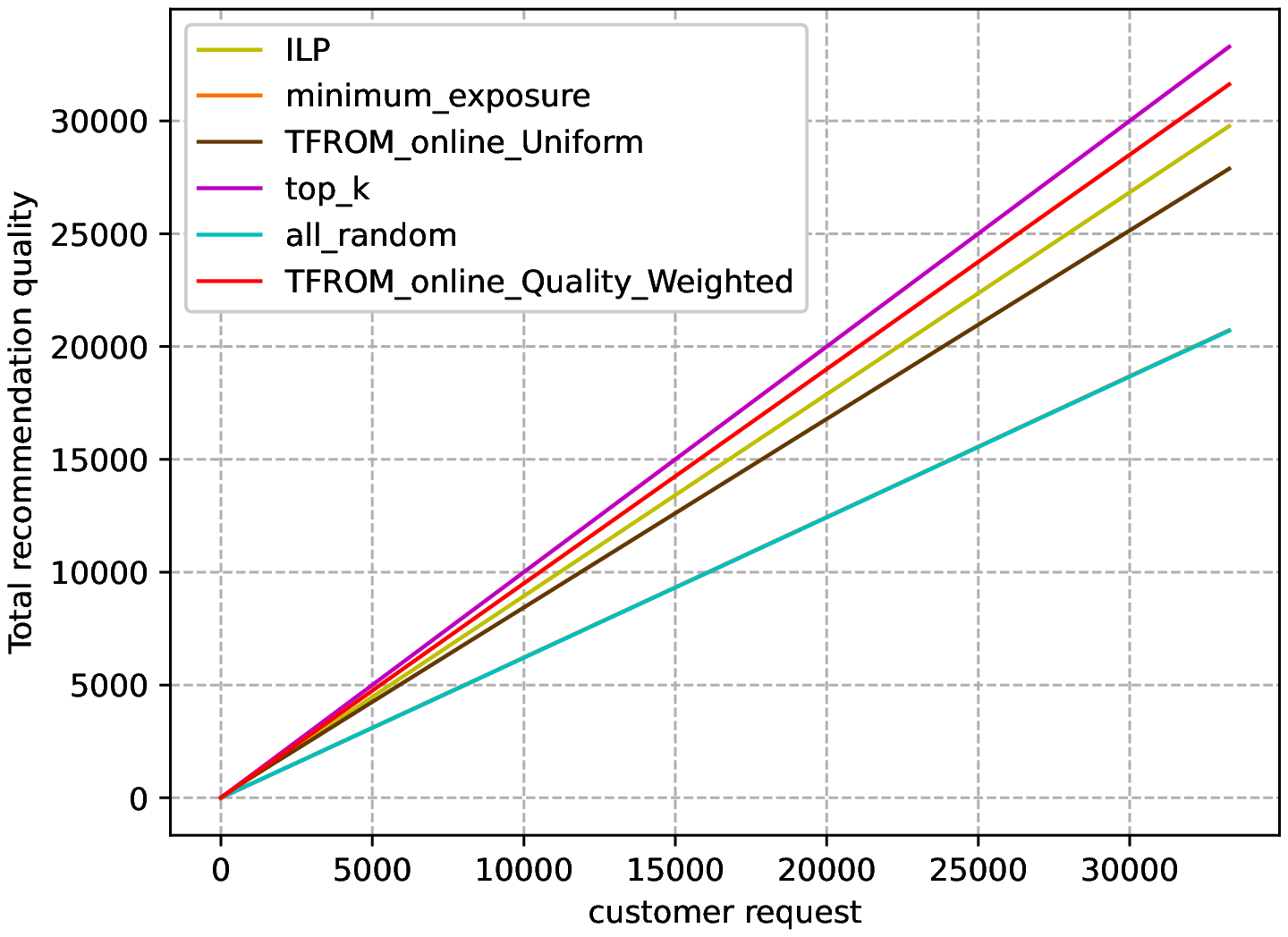}
			\end{minipage}%
		}%
		\subfigure[Variance of NDCG]{
			\begin{minipage}[t]{0.25\linewidth}
				\centering
				\includegraphics[width=\textwidth,height=2.5cm]{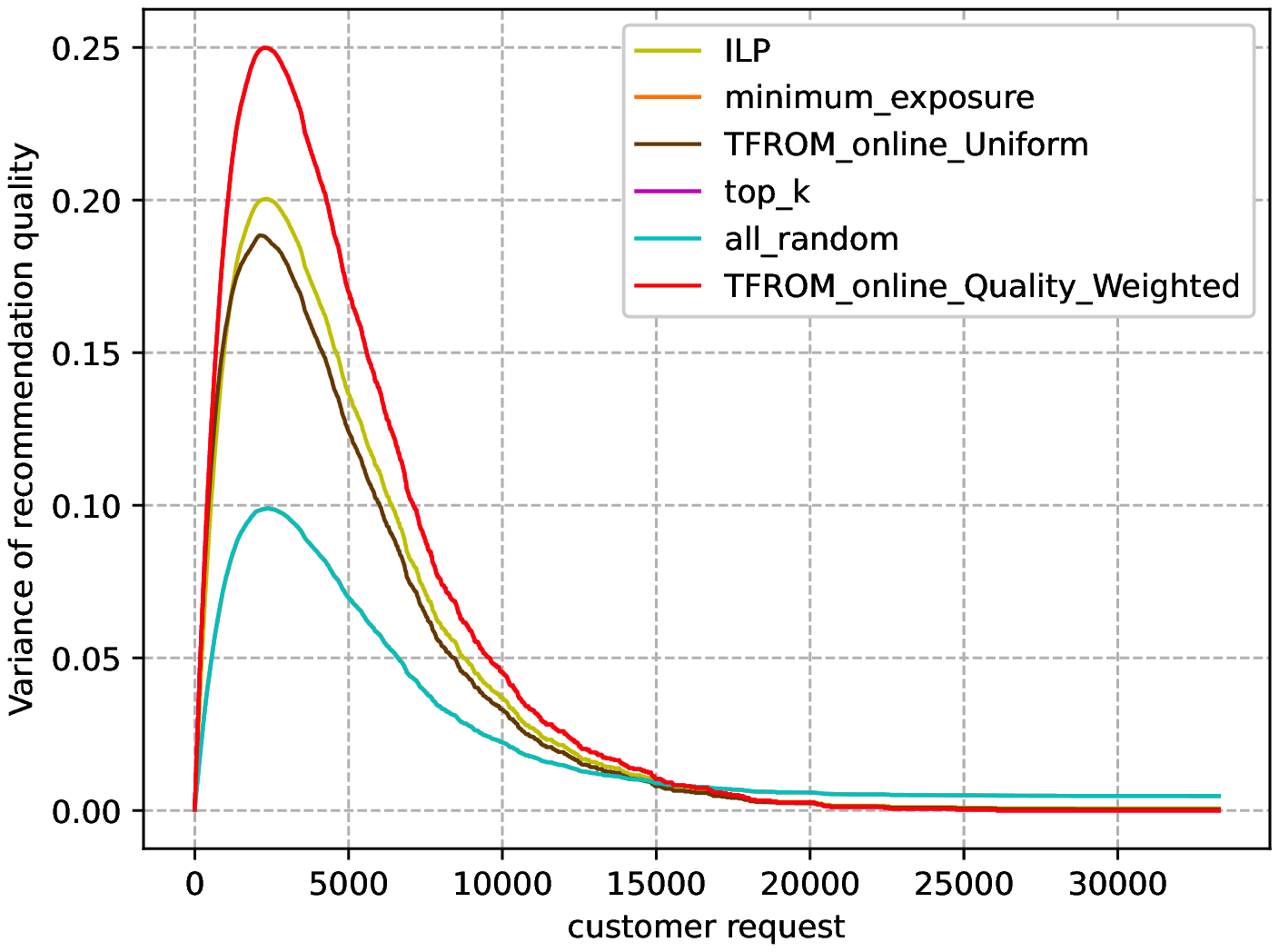}
			\end{minipage}
		}%
		\subfigure[Variance of exposure]{
			\begin{minipage}[t]{0.25\linewidth}
				\centering
				\includegraphics[width=\textwidth,height=2.5cm]{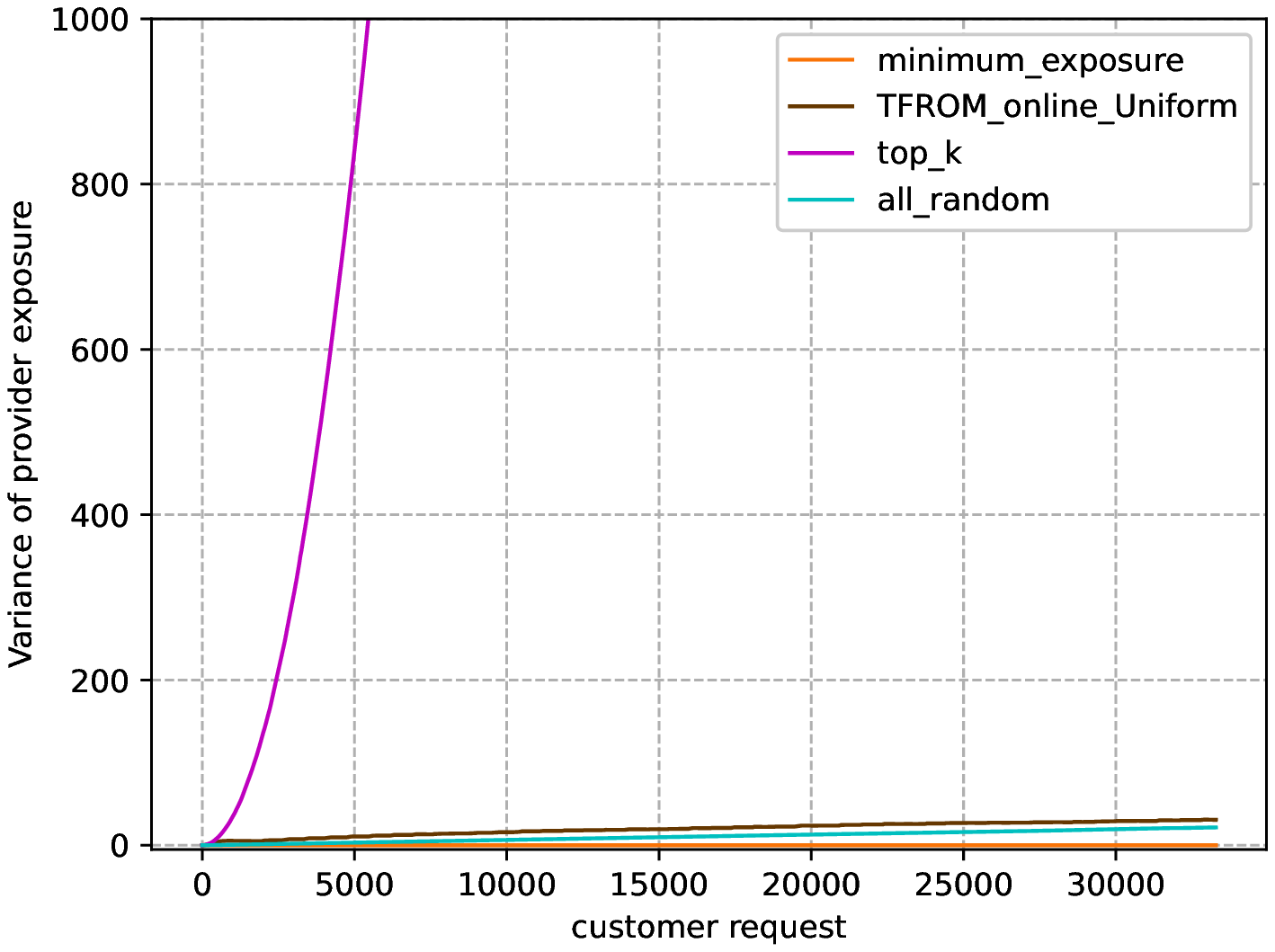}
			\end{minipage}
		}%
		\subfigure[Variance of the ratio of exposure and relevance]{
			\begin{minipage}[t]{0.25\linewidth}
				\centering
				\includegraphics[width=\textwidth,height=2.5cm]{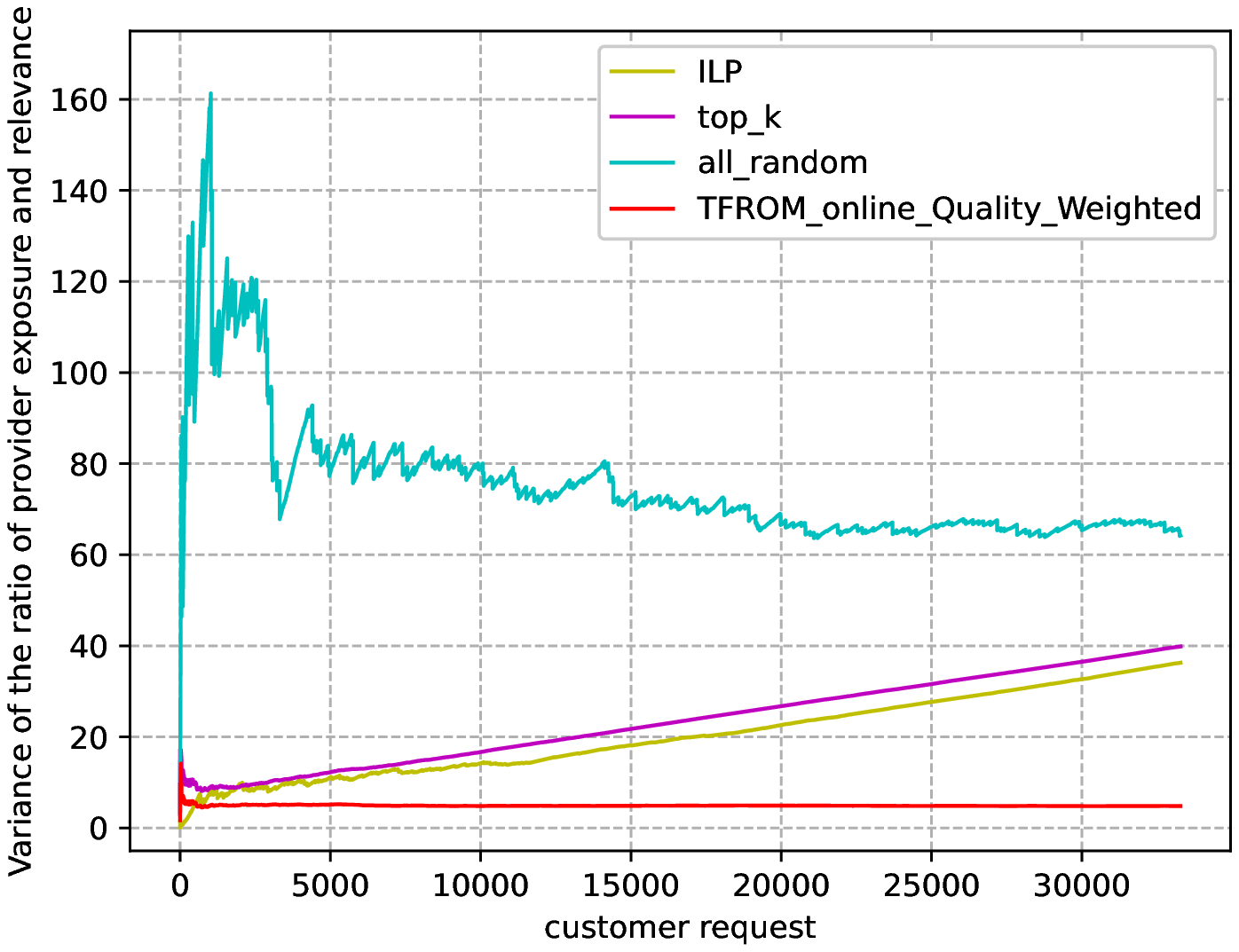}
			\end{minipage}
		}%
		\centering
		\caption{Experiment Results on Google Dataset in the Online Scenario}
		\label{fig6}
	\end{figure*}
	
	\begin{figure*}[!h]
		\centering
		\subfigure[Total recommendation quality]{
			\begin{minipage}[t]{0.25\linewidth}
				\centering
				\includegraphics[width=\textwidth,height=2.5cm]{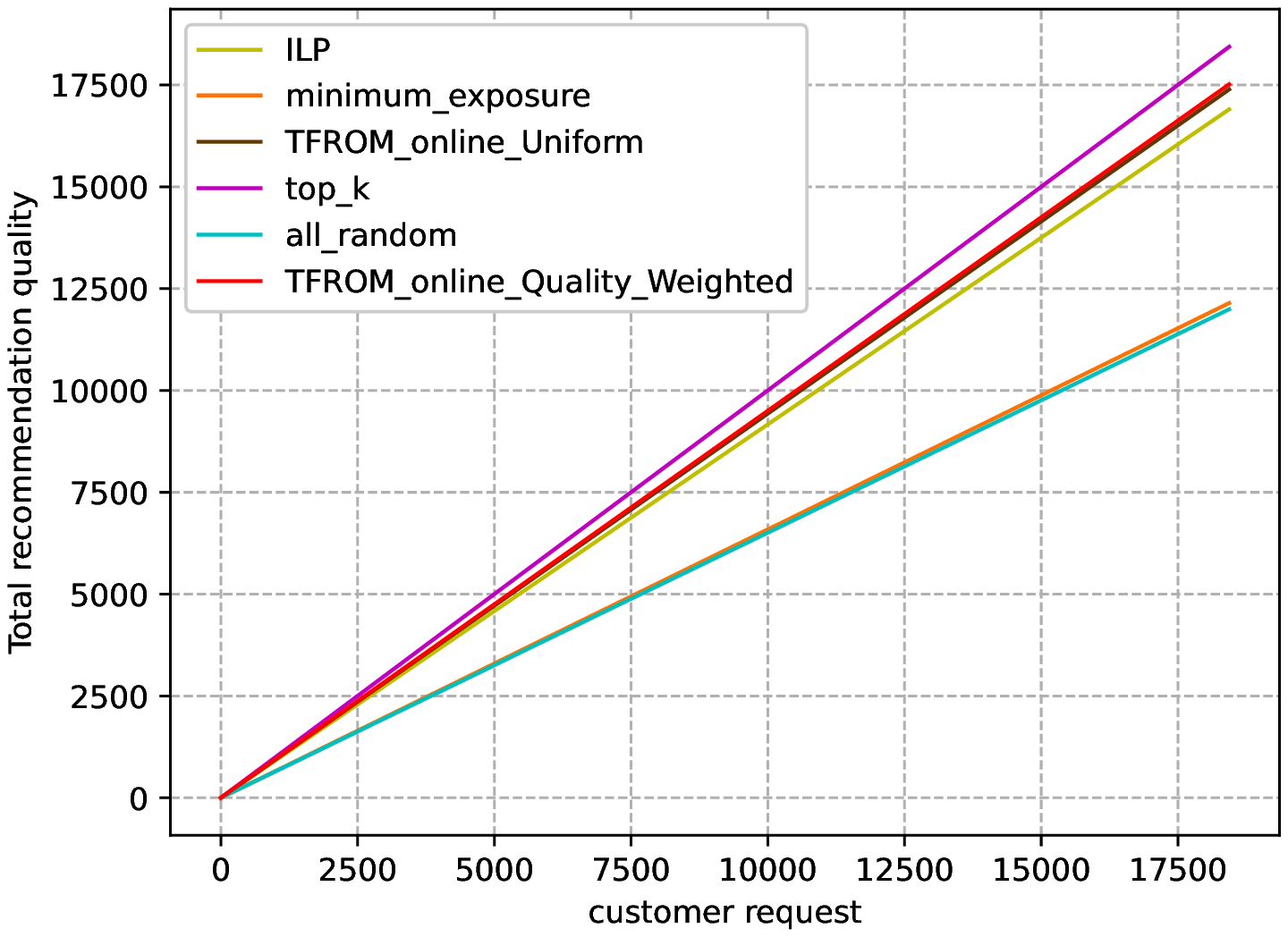}
			\end{minipage}%
		}%
		\subfigure[Variance of NDCG]{
			\begin{minipage}[t]{0.25\linewidth}
				\centering
				\includegraphics[width=\textwidth,height=2.5cm]{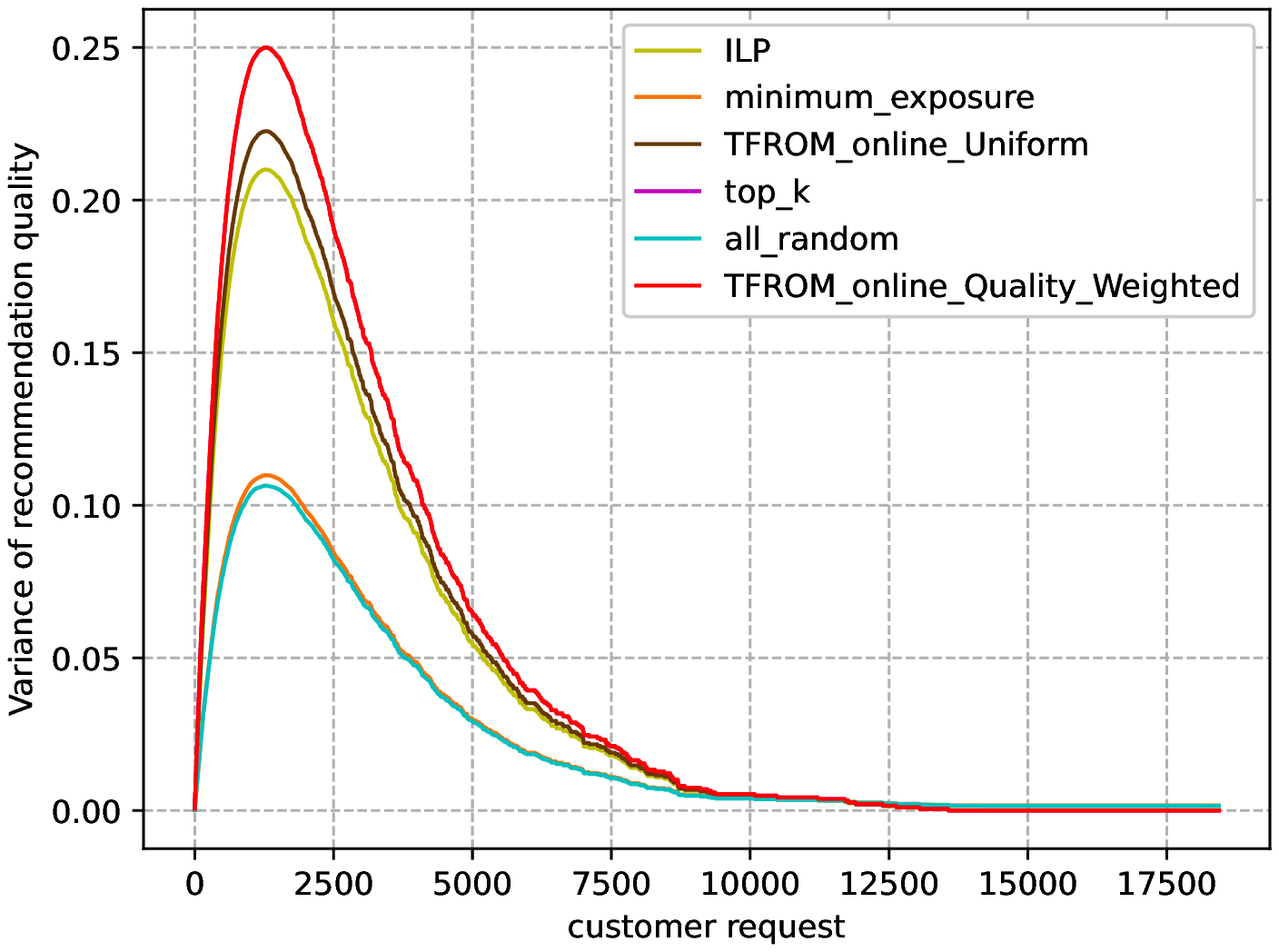}
			\end{minipage}
		}%
		\subfigure[Variance of exposure]{
			\begin{minipage}[t]{0.25\linewidth}
				\centering
				\includegraphics[width=\textwidth,height=2.5cm]{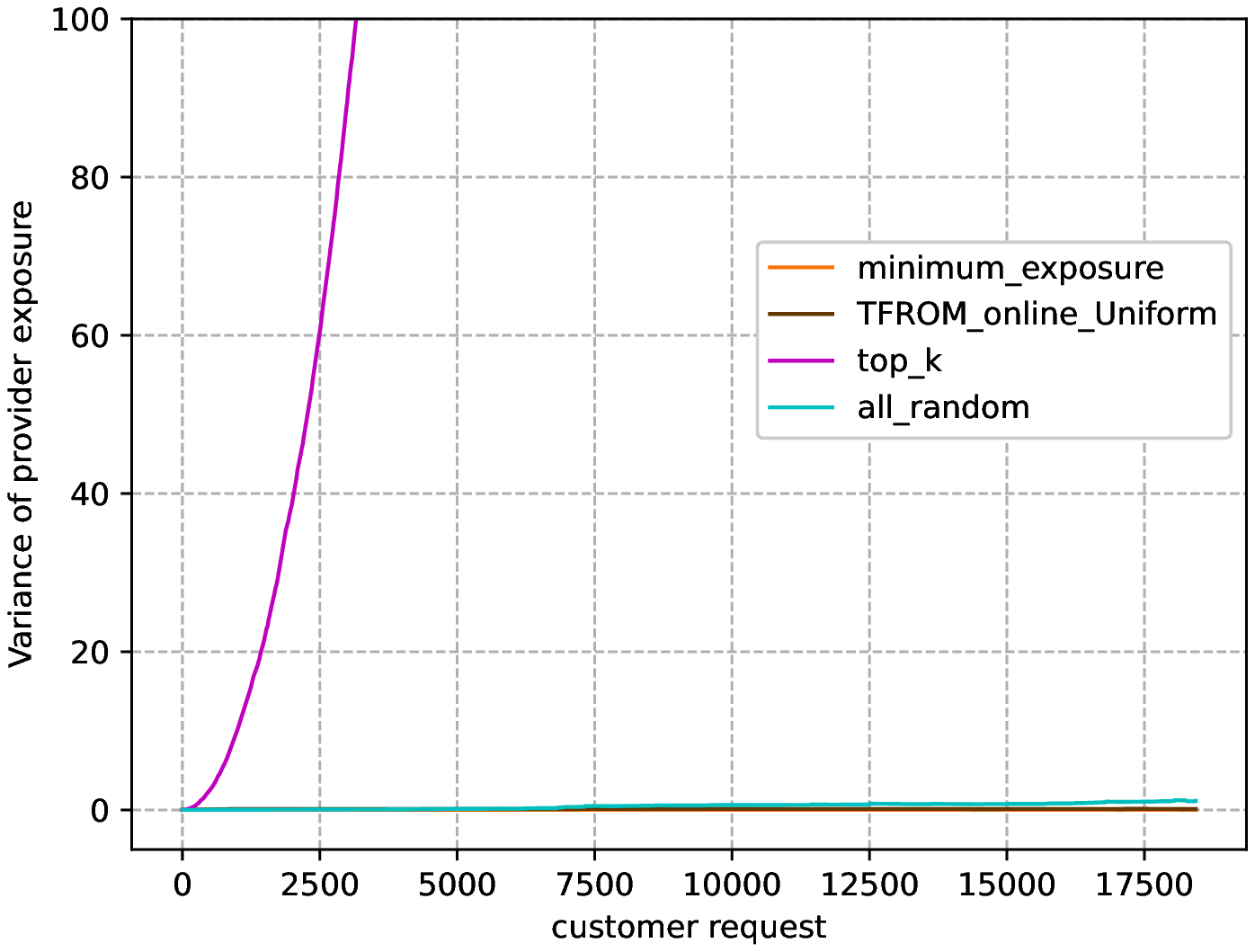}
			\end{minipage}
		}%
		\subfigure[Variance of the ratio of exposure and relevance]{
			\begin{minipage}[t]{0.25\linewidth}
				\centering
				\includegraphics[width=\textwidth,height=2.5cm]{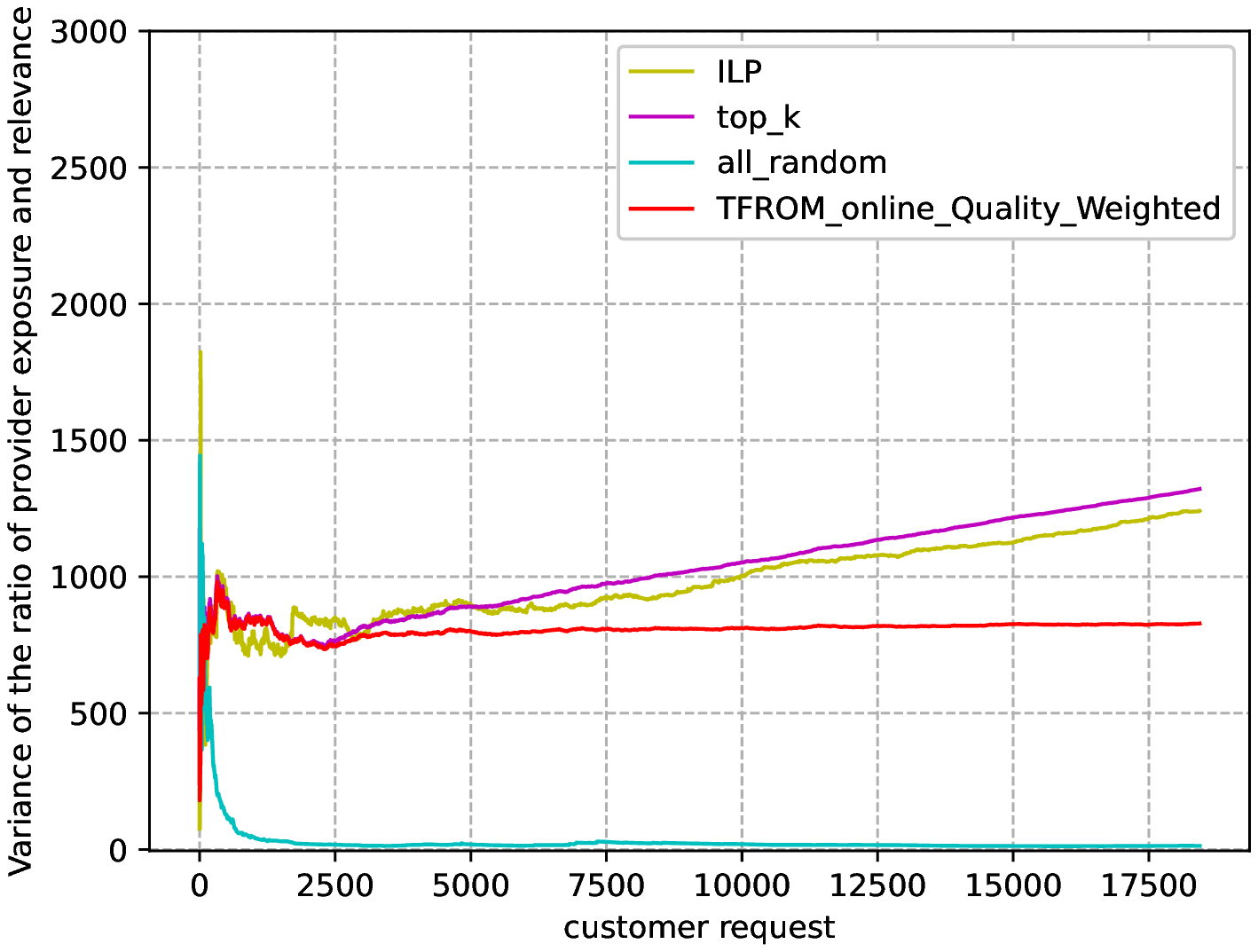}
			\end{minipage}
		}%
		\centering
		\caption{Experiment Results on Amazon Dataset in the Online Scenario}
		\label{fig7}
	\end{figure*}

	\subsubsection{Results for the online situation}
	We generate a random sequence of customer requests to simulate the online scenario, and evaluate the changes of the aforementioned metrics during the recommendation process. The length of the sequence is set to 10 times the number of customers, so that each customer can receive multiple recommendations. At the same time, we also test the performance of  \textit{TFROM-online} with two kinds of provider-side fairness. Since \textit{FairRec} is not suitable for online scenarios, we do not compare it in this experiment.  
	
	\paragraph{Recommendation quality}
	It can be seen from Figures \ref{fig5}(a), \ref{fig6}(a) and \ref{fig7}(a) that with the continuous arrival of customer requests, the recommendation quality of the aforementioned algorithms grows basically linearly. \textit{Top-k} still achieves the maximum value for recommendation quality, whereas the \textit{Minimum exposure} algorithm achieves the worst result by completely ignoring recommendation quality. \textit{TFROM-online} has a small amount of loss on the three data sets. Of these, the loss of \textit{TFROM-online-Uniform} is larger because the goal of uniform fairness forces the algorithm to select items with lower relevance in the recommendation list of a single customer. This loss is accumulated during the recommendation process, but in fact, for a single recommendation, the loss of quality is completely acceptable to a customer.
	
	\paragraph{Customer-side Fairness}
	The results are shown in Figures \ref{fig5}(b), \ref{fig6}(b) and \ref{fig7}(b). It is worth noting that customer fairness decreases firstly and then increases with  the number of requests (variance and mean deviation increases first and then decreases). This is because at the beginning, most customers have not received recommendation, and the customer's recommendation quality changes from zero to a value greater than zero. This change is dramatic, resulting in the rapid improvement of the deviation of recommendation quality. After all the customers have received at least one recommendation, the subsequent recommendation will reduce the differences between customers and the system tends to be stable. Compared with other algorithms, \textit{TFROM-online} can reach a good level of customer fairness in the end which shows that \textit{TFROM-online} can maintain the fairness of customer recommendation quality in the long-term recommendation process.
	
	\paragraph{Provider-side Fairness}
	As shown in Figures \ref{fig5}(c), \ref{fig6}(c) and \ref{fig7}(c), if no action is taken on exposure unfairness as in the \textit{Top-k} algorithm, the degree of unfairness will continue to increase along with the recommendation process and will reach an unacceptable level. At the same time, the other algorithms can provide very good exposure fairness, which is consistent with the results of offline scenarios. As for \textit{Quality Weighted Fairness}, the experiment results are basically the same as the offline scenarios. Compared with the comparison algorithms, \textit{TFROM-online-Quality-Weighted} can provide recommendation results more consistently and more fairly.

	\section{Conclusions}
	In this paper, we consider the issue of fairness in a recommendation system from two sides, i.e., customers and providers. The objective of our study is to ensure fairness for both sides while maintaining a high level of personalization in the recommendation results. We model the providers that provide multiple items and ensure fairness among them at the group level and consider two kinds of fairness definitions, while from the customer perspective, we ensure fairness between individual customers. Aiming at both offline and online scenarios, we design post-processing heuristic algorithms to ensure two-sided fairness, which enables our method to be easily applied to various existing recommendation systems in various scenarios, and helps them to improve the fairness of the system. Experiments on three real-world datasets show that our algorithms provide better two-sided fairness than the comparison algorithms while losing only a little recommendation quality.
	
	\bibliographystyle{ACM-Reference-Format}
	\bibliography{ref}
	
\end{document}